\documentclass{article} 
\usepackage{iclr2026_conference,times}

\usepackage{amsmath,amsfonts,bm}

\def\eqref#1{equation~\ref{#1}}

\def\1{\bm{1}}

\DeclareMathAlphabet{\mathsfit}{\encodingdefault}{\sfdefault}{m}{sl}
\SetMathAlphabet{\mathsfit}{bold}{\encodingdefault}{\sfdefault}{bx}{n}

\newcommand{\R}{\mathbb{R}}

\usepackage{hyperref}
\usepackage{url}

\usepackage{graphicx} 
\usepackage{natbib}  
\usepackage{caption} 
\usepackage{amsmath} 
\usepackage{amssymb} 
\usepackage{booktabs} 
\usepackage{multirow} 
\usepackage{cleveref} 
\usepackage{tabularx}
\usepackage{algorithm}
\usepackage{algorithmic}
\usepackage{subcaption}
\usepackage{wrapfig}
\usepackage{caption}
\usepackage{float}
\title{Expressive yet Efficient Feature Expansion \\with Adaptive Cross-Hadamard Products}

\author{Xuyang Zhang\textsuperscript{1,2}, Xi Zhang\textsuperscript{2}, Liang Chen\textsuperscript{1,2}\thanks{Corresponding author.\ (\texttt{chenl@bit.edu.cn})}, Hao Shi\textsuperscript{1,2}, Qingshan Guo\textsuperscript{2}\\
\textsuperscript{1}Beijing Institute of Technology, Beijing, China\\
\textsuperscript{2}Chongqing Innovation Center, 
Beijing Institute of Technology, Chongqing, China
}

\iclrfinalcopy 
\begin{document}

\maketitle

\begin{abstract}
Recent theoretical advances reveal that the Hadamard 
product induces nonlinear representations and implicit 
high-dimensional mappings for the field of deep learning, 
yet their practical deployment in resource-constrained vision models 
remains largely unexplored. To address this gap, we introduce 
the Adaptive Cross-Hadamard (ACH) module, a novel operator 
that embeds learnability through differentiable discrete 
sampling and dynamic softsign normalization. This facilitates highly 
efficient feature reuse without incurring additional convolutional 
parameters, while ensuring stable gradient flow. 
Integrated into Hadaptive-Net 
(Hadamard Adaptive Network) via neural 
architecture search, our approach achieves unprecedented 
efficiency. Comprehensive experiments demonstrate 
state-of-the-art accuracy/speed trade-offs on image 
classification tasks, establishing Hadamard operations as 
specific building blocks for efficient vision models.
The source code is available at 
\url{https://github.com/acelych/hadaptivenet}.
\end{abstract}

\section{Introduction}
\label{sec:intro}

Since AlexNet revolutionized computer vision 
\citep{krizhevsky2012imagenet}, deep convolutional neural 
networks (CNNs) have advanced rapidly. Subsequent 
innovations mitigated gradient explosion via residual 
connections \citep{he2016deep} and integrated self-attention 
into vision architectures \citep{dosovitskiy2020image}, 
gradually shifting model design toward greater depth for 
performance gains.

Conversely, lightweight networks 
\citep{howard2017mobilenets,zhang2018shufflenet,
ma2018shufflenet,han2020ghostnet} pursued efficiency. 
These models widely adopted the inverted bottleneck structure 
(e.g., MobileNets \citep{howard2017mobilenets,
sandler2018mobilenetv2,howard2019searching,qin2024mobilenetv4}, 
ConvNext \citep{liu2022convnet,woo2023convnext}), which expands 
channel dimensions within blocks rather than compressing them. 
This design enables residual operations in lower-dimensional 
spaces, reducing computation while mitigating representational 
redundancy in high dimensions. 

However, the inverted residual structure’s dependency on 
repeated channel expansion/reduction operators inevitably 
introduces computational redundancy. Although effective, its 
dimension expansion phase requires significant convolution 
operations to project features into high-dimensional spaces, 
where substantial similarity exists across newly generated 
channels. GhostNet \citep{han2020ghostnet,Tang2022GhostNetV2EC,
Liu2024GhostNetV3ET} reveals this critical inefficiency, 
demonstrating that a large portion of expanded channels exhibit 
high linear correlations, and thus can be inexpensively 
synthesized via learned linear transformations of primary 
features rather than redundant convolutions. This breakthrough 
established the first generalized framework for feature reuse, 
bypassing costly dimension-specific operations.
Subsequent works like FasterNet \citep{Chen_2023_CVPR} further 
refined this paradigm, implementing feature reuse via partial 
convolution operators that selectively merge spatially 
neighboring features. 

\begin{figure}[t]
\begin{subfigure}[t]{0.49\textwidth}
  \centering
  \includegraphics[width=\textwidth]{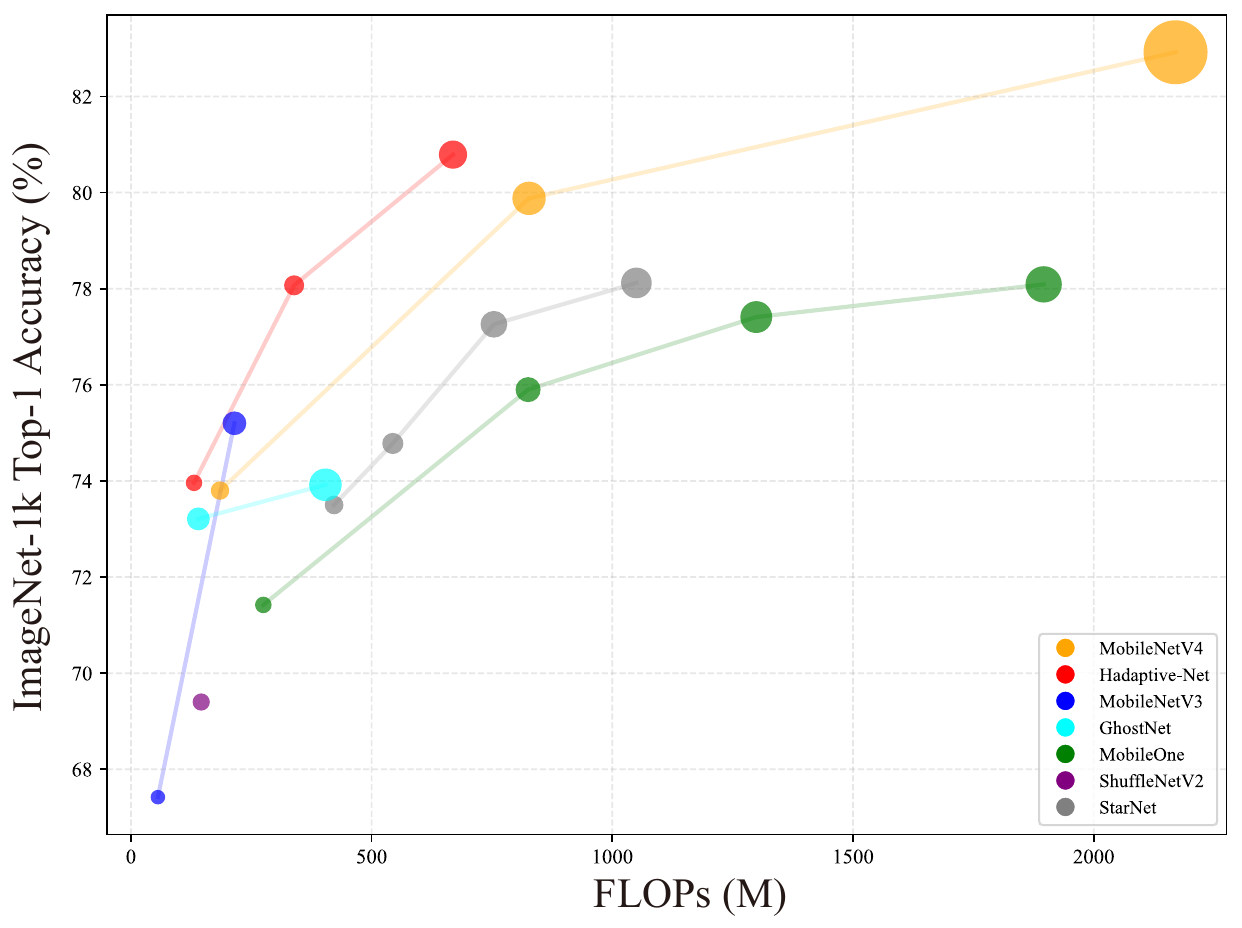}
\end{subfigure}
\hfill
\begin{subfigure}[t]{0.49\textwidth}
  \centering
  \includegraphics[width=\textwidth]{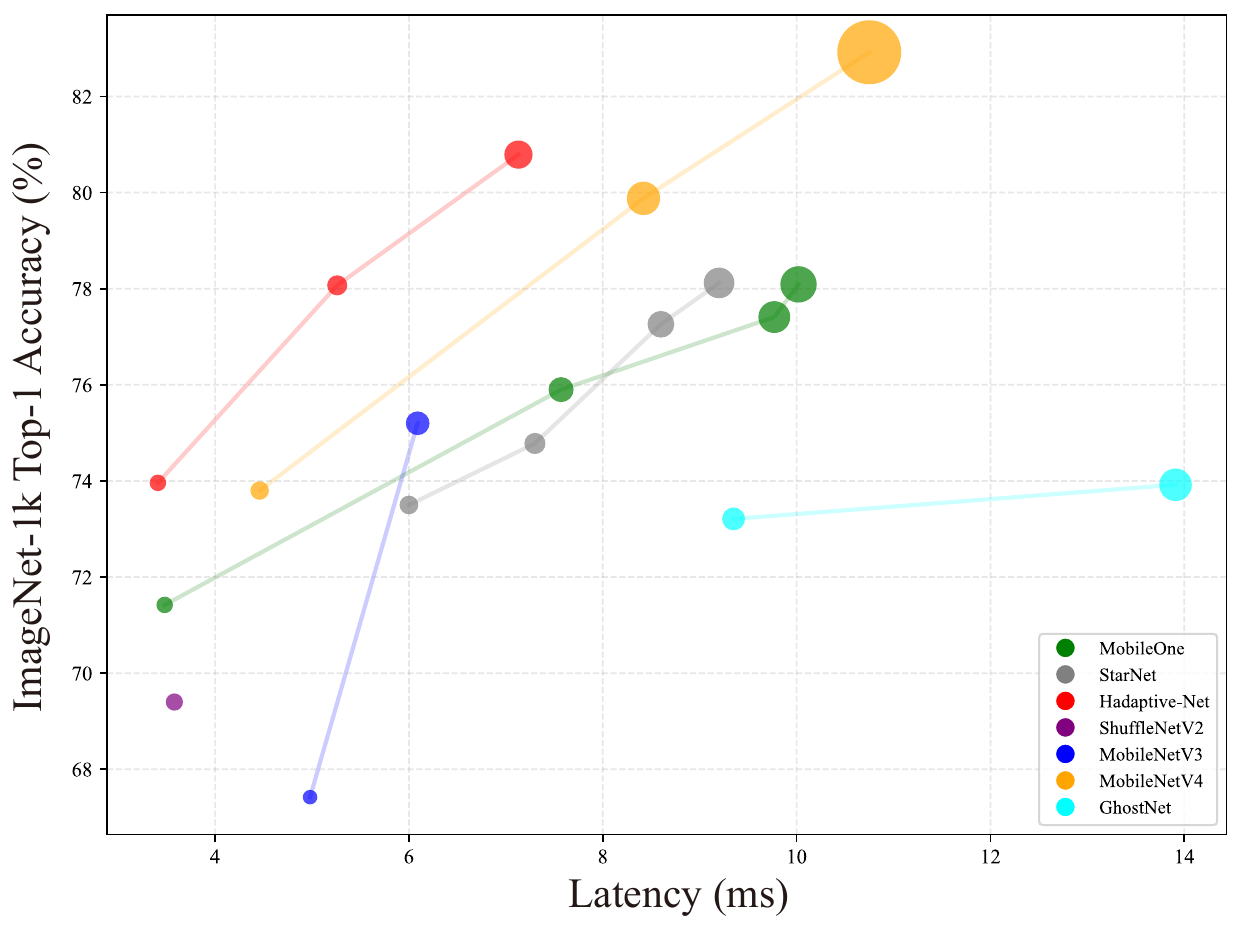}
\end{subfigure}
\caption{
   \textbf{The trade-off between FLOPs/latency and top-1 accuracy.}
   These diagrams compare the efficiency among different state-of-the-art 
   models with ours Hadaptive-Net in image classification task. 
   Detailed experimental configurations are provided in 
   \cref{subsec:img-class}.
}
\label{fig:0}
\end{figure}

Our work revisits feature recombination efficiency from an 
orthogonal perspective: instead of generating or filtering 
features, we exploit the intrinsic nonlinear representational 
capacity of learnable Hadamard products to achieve 
ultra-efficient feature fusion. 
The Hadamard product (a.k.a. element-wise multiplication), 
as a highly practical method, has long garnered significant 
attention in the fields of deep learning. 
Recently, it became a new learning paradigm in the field of 
lightweight network design owing to 
effective performance and concise computation. 
Its principle is straightforward, 
for two identical matrices $\mathbf{A},\mathbf{B}$:
\[
   \mathbf{C}=\mathbf{A}\odot\mathbf{B}\Leftrightarrow 
   C_{i,j}=A_{i,j}\cdot B_{i,j}
\]
Recent theoretical advances reveal that stacked Hadamard 
products can induce nonlinear representations and implicitly 
high-dimensional mappings when deeply cascaded 
\citep{ma2024rewrite}. Capitalizing on these insights, we 
propose the Adaptive Cross-Hadamard (ACH) module. This novel 
operator transcends conventional Hadamard usage by embedding 
learnability through two key mechanisms: (i) channel 
attention-guided feature gating, and (ii) differentiable 
discrete sampling. Thus, ACH establishes 
Hadamard products as foundational deep learning operators while 
enabling parameter-free feature reuse.

To effectively deploy the ACH module, we construct Hadaptive-Net 
(Hadamard Adaptive Network) through differentiable neural 
architecture search (NAS), jointly optimizing model topology 
and ACH integration points. For efficient on-device execution, 
we further develop tailored GPU acceleration strategies 
addressing computation scheduling challenges. In comparative 
experiments, Hadaptive-Net outperforms state-of-the-art 
efficient models, achieving higher accuracy with lower 
computational costs (\cref{fig:0}).

\section{Related Work}
\label{sec:related}

This section reviews two types of previous studies related 
to this work: the application of Hadamard product and 
efficient model design.

\subsection{Researches in Hadamard}

It can be learned from 
\citet{DBLP:journals/pami/ChrysosWPTC25} that the taxonomy for 
applying the Hadamard product in deep learning is divided into 
four categories: high-order interactions, multimodal fusion, 
adaptive modulation, and efficient operators. 
\citet{ma2024rewrite} and \citet{chen2022simple} reveal 
its ability to implicitly induce high-order nonlinear mappings. 
As an example of multimodal fusion, 
\citet{DBLP:conf/iclr/KimOLKHZ17} uses the Hadamard product to 
achieve low-rank bilinear pooling as an approximation of full 
bilinear pooling. Adaptive modulation—also referred to as the 
gating mechanism, such as in LSTMs \citep{hochreiter1997long}
—is a widely adopted application of the Hadamard product. 
For instance, HAda \citep{wang2024hada} employs it to scale 
weights generated by a hypernetwork in multi-view learning 
scenarios, while HiRA \citep{huanghira} applies it to construct 
high-rank weight updates during the fine-tuning of large language 
models. MogaNet \citep{DBLP:conf/iclr/LiW00L00ZL24} also uses the 
Hadamard product to adaptively focus on informative features by 
fusing multi-scale depthwise separable convolutions with 
varying dilation rates.

The forms of efficient operators are quite diverse. 
To mitigate the $\mathcal{O}(n^2)$ complexity of Transformers, 
some approaches replace matrix multiplications in attention 
mechanisms with Hadamard products, as seen in 
FocalNet \citep{yang2022focal} and HorNet \citep{rao2022hornet}. 
\citet{DBLP:journals/corr/abs-2312-00752} and 
\citet{DBLP:conf/icml/ZhuL0W0W24} adopt the Hadamard product as 
a core operator for ultra-efficient feature expansion and 
nonlinear fusion via channel-wise cross-products. However, 
existing methods suffer from critical limitations: 
fixed combination rules (inter- or intra-channel) restrict 
optimization flexibility, and predefined operations limit 
interpretability. We therefore propose enhancing Hadamard 
products with learnable channel expansion capabilities, 
transforming them into dedicated deep learning operators 
that leverage inherent nonlinearity while overcoming previous 
rigidity.



\subsection{Efficient Model Design}

The pursuit of efficient architectures has driven continuous 
innovation: from SqueezeNet's pioneering use of pointwise 
convolutions \citep{Iandola2016SqueezeNetAA}, to MobileNetV1's 
depthwise separables \citep{howard2017mobilenets}, MobileNetV2's 
inverted bottlenecks \citep{sandler2018mobilenetv2}, and 
ShuffleNet's channel shuffling 
\citep{zhang2018shufflenet,ma2018shufflenet}. 
Neural Architecture Search (NAS) further advanced efficiency 
in MnasNet \citep{tan2019mnasnet}, EfficientNet 
\citep{tan2019efficientnet}, and MobileNetV3 
\citep{howard2019searching}, culminating in MobileNetV4's 
universal inverted bottlenecks \citep{qin2024mobilenetv4}. 
Concurrently, vision transformers inspired hybrid designs like
Mobile-Former \citep{Chen_2022_CVPR} and EdgeViT 
\citep{Pan2022EdgeViTsCL}.

Feature reuse mechanisms provide complementary efficiency: 
GhostNet revealed channel-wise redundancies in conventional 
convolutions, replacing redundant features via linear 
transformations \citep{han2020ghostnet,Tang2022GhostNetV2EC}. 
FasterNet constrained convolution ranges \citep{Chen_2023_CVPR}, 
while GhostNetV3 \citep{Liu2024GhostNetV3ET} and MobileOne 
\citep{AnasosaluVasu2022MobileOneAI} adopted RepVGG's 
reparameterization \citep{Ding2021RepVGGMV} to merge parallel 
branches.

\begin{figure*}[!t]
   \centering
   \includegraphics[width=\textwidth]{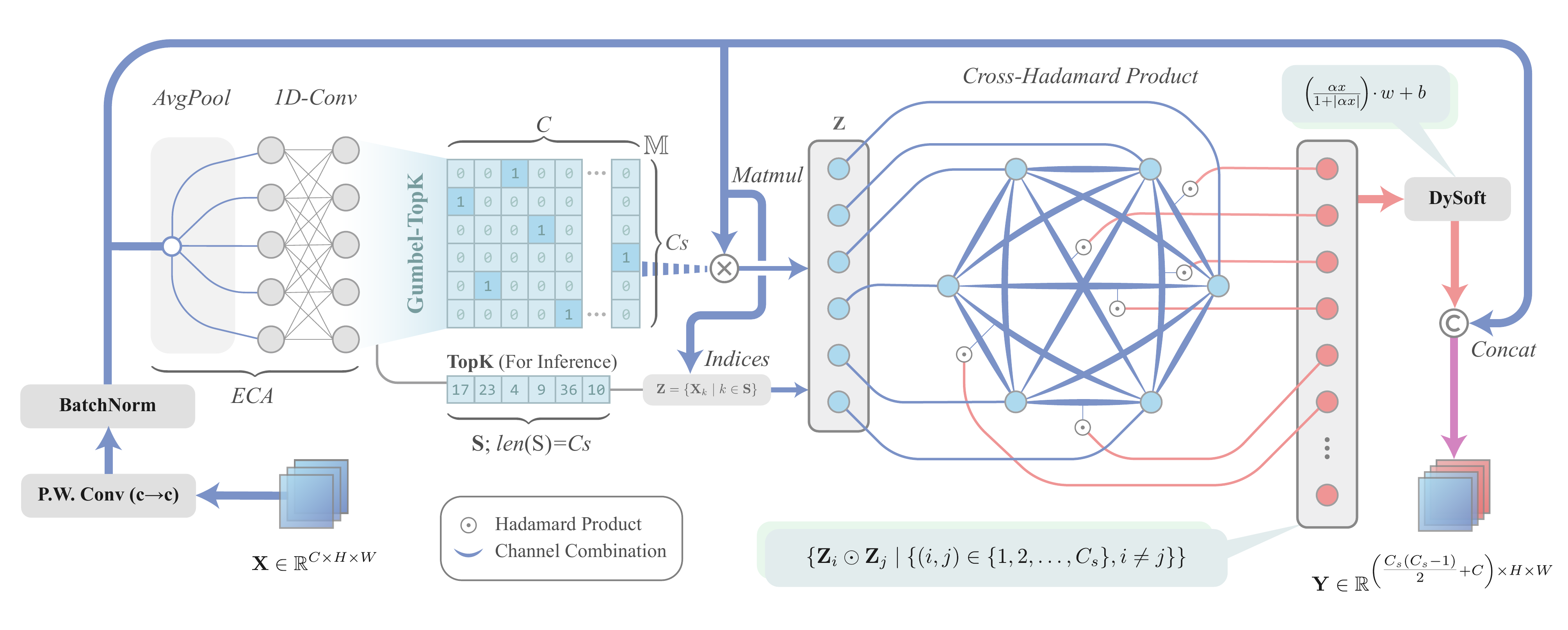}
   \caption{
      \textbf{Illustration of the ACH module.}
      Input features $\mathbf{X}$ undergo linear transformation 
      and batch normalization. An ECA module generates 
      channel-wise scores, with Gumbel-Topk sampling (training) 
      or top-k selection (inference) determining active channels. 
      Selected features $\mathbf{Z}$ undergo cross-Hadamard 
      product, normalized by dynamic softsign, 
      then concatenated with original features.
   }
   \label{fig:2}
\end{figure*}

\section{Methodology}
\label{sec:method}

This section establishes a hierarchical framework for the 
ACH module, progressing from mathematical foundations to 
architectural deployment. First, we formalize the Hadamard 
product's role in channel expansion. Second, we introduce 
differentiable discrete sampling via Gumbel-TopK with adaptive 
temperature annealing, enabling end-to-end channel selection. 
Third, to stabilize dynamically generated features, 
DySoft normalization replaces statistical normalization with 
bounded sigmoidal activation. Finally, we integrate ACH into 
Hadaptive-Net through gradient-based NAS.

\subsection{Hadamard for Channel Expansion}

Inspired by the properties of high-dimensional mapping and 
non-linearity, we observe that the Hadamard product aligns well 
with the characteristic of neural networks that 
gradually increase channel dimensions 
while reducing spatial dimensions. 
This suggests that the Hadamard product is particularly 
suitable for channel expansion.

Specifically, we compute the Hadamard product for 
pairwise combinations of input channels while retaining 
the original feature maps. This can be expressed as:

\begin{equation}\label{eq:1}
   \begin{array}{l}
      \mathbf{Y}=
      \mathbf{X}\oplus
      \{
         \mathbf{X}_i\odot\mathbf{X}_j
         \mid
         \{(i,j)\in\{1,2,\dots,C\}, i\neq j\}
      \}
      \\[4pt]
      \text{s.t.}\quad
      \mathbf{X}\in\mathbb{R}^{C\times H\times W},
      \mathbf{Y}\in\mathbb{R}^{\tfrac{C(C+1)}{2}\times H\times W}
   \end{array}
\end{equation}

\noindent where $\mathbf{X}$ represents the input feature map,
$\mathbf{X_i}$ and $\mathbf{X_j}$ denote the $i$-th 
and $j$-th channels of $\mathbf{X}$, $\odot$ denotes 
Hadamard product, and $\oplus$ denotes channel-wise 
concatenation, respectively.
This approach can be seen as putting the initial features 
$\mathbf{X}$ and the features after transformation into 
the same feature space. More specifically, 
the stitched feature vector can be understood as a 
high-dimensional vector, and the original feature space can be 
regarded as a set of bases, providing interpretability for the 
composite features that carry 
implicit high-dimensional information. 

Based on these insights, we designed the Adaptive Cross-Hadamard 
module, which is illustrated as \cref{fig:2}. 
The design details and learnable methods of the module will be 
discussed in the following sections.

\subsection{Differentiable Discrete Sampling}

As feature maps propagate through deep networks, their 
channel dimensions expand dramatically, causing the number of 
possible channel interactions to grow quadratically. This 
combinatorial explosion makes exhaustive pairwise computation 
prohibitively expensive. Even for modest channel counts, 
practical implementations require selecting a fixed subset of 
channels for efficient processing. 
We thus reformulate \cref{eq:1} as:

\begin{equation}\label{eq:2}
   \begin{array}[pos]{l}
      \mathbf{Y}=
      \mathbf{X}\oplus
      \{
         \mathbf{Z}_i\odot\mathbf{Z}_j
         \mid
         \{(i,j)\in\{1,2,\dots,C_{(\text{s})}\}, i\neq j\}
      \}
      \\[4pt]
      \text{s.t.}\quad
      \mathbf{X}\in\mathbb{R}^{C\times H\times W},
      \mathbf{Y}\in\mathbb{R}^{\left(\tfrac{C_{(\text{s})}(C_{(\text{s})}-1)}{2}+C\right)\times H\times W}
      \\[4pt]
      \text{s.t.}\quad
      \mathbf{Z}=\{\mathbf{X}_k\mid k\in\mathbf{S}\}
   \end{array}
\end{equation}

\noindent where $\mathbf{S}$ represents a sequence 
of chosen channels' indexes and $C_{(\text{s})}$ indicates the amount of 
chosen channels.

However, this selection process is inherently discrete, 
posing a challenge for gradient-based optimization.  
Thus, we introduced Gumbel-Topk trick
 \citep{gumbel1954statistical} for selecting procedure.
Formally, we donate scores of each channels as a vector $\xi$, 
which is obtain from an ECA module \citep{Wang_2020_CVPR}:

\begin{equation}
    \xi=\operatorname{ECA}(\mathbf{X})=\mathcal{P}(\mathbf{X})\ast W + b
\end{equation}

\noindent where $\mathcal{P}$ denotes adaptive average pooling, 
$\ast$ denotes a 1D convolution operation. 
Then calculate the probability distribution as below:

\begin{equation}\label{eq:3}
   \begin{array}{l}
      \mathbf{M}_c = \dfrac
      {\exp{\left(\dfrac{\xi_c + o_c}{\tau}\right)}}
      {
         \sum_{c'=1}^{C}
         {\exp{\left(\dfrac{\xi_{c'} + o_{c'}}{\tau}\right)}}
      }
      \quad c\in C
      \\[20pt]
      \text{s.t.}\quad
      o_i = -\log(-\log(u)),\ u\sim\text{Unif}\ [0,1]
   \end{array}
\end{equation}

\noindent where $o_i$ are i.i.d sampled from Gumbel distribution,
$\mathbf{M}$ denotes a probability distribution vector 
resulted from softmax, and $\tau$ denotes temperature 
parameter that controls the smoothness of the softmax output, 
respectively. The Gumbel-distributed perturbations $o_i$ inject 
controlled stochasticity into the discrete selection process, 
ensuring channels temporarily still receive gradient feedback. 
This prevents over-reliance on initial channel selections while 
maintaining alignment with the ECA's distribution across 
forward passes. The temperature parameter $\tau$ governs 
output sharpness: higher values yield softer selections, 
while $\tau\to0$ produces one-hot behavior.

While $\mathbf{M}$ is continuous and differentiable, 
it leads to a discrete and nondifferentiable 
vector $\mathbf{M^H}$. With straight through estimator (STE) 
technique \citep{bengio2013estimating}:

\begin{equation}\label{eq:4-1}
   \quad\mathbf{M}^\text{H}_c = 
   I_{\mathbf{S}}(c),\ 
   \mathbf{S}=\operatorname{top-k}(\mathbf{M}, k=C_{(\text{s})})
   \quad c\in [1,C] \quad
   \text{forward}
\end{equation}
\begin{equation}\label{eq:4-2}
   \frac{\partial \mathcal{L}}{\partial \xi} 
   :=\frac{\partial \mathcal{L}}{\partial \mathbf{M}^\text{H}}\cdot\frac{\partial \mathbf{M}}{\partial \xi}
   =\frac{\partial \mathcal{L}}{\partial \mathbf{M}^\text{H}}\cdot\frac{\partial \operatorname{softmax}(\xi/\tau)}{\partial \xi}
   \quad\text{backward}
\end{equation}

\noindent where $I_A(x)$ denotes the indicator function, 
discrete $\mathbf{M^H}$ could conduct data stream during 
training and gradient could skip through 
$\mathbf{M^H}$ to $\mathbf{M}$ during backpropagation. 
Since hyperparameter $\tau$ modulates the intensity of continuous values 
influence the selection of discrete values through softmax, 
Consequently, the adjustment of $\tau$ should be responsive to 
gradient variations. Instead of relying on a global parameter 
scheduler, the ACH module employs adapting $\tau$ dynamically 
based on the norm of historical gradients, 
thereby preserving the end-to-end training characteristics:

\begin{equation}\label{eq:tau-adj}
   \begin{array}{rcl}
      \tau &\leftarrow& \text{CLAMP}\left( \tau \cdot \left(1 + \alpha \cdot \text{sign}(\|grad\|_2 - \tau_{hist})\right),\ 0.01,\ 4.0 \right)\\
      \tau_{hist} &\leftarrow& \|grad\|_2 \\
   \end{array}
\end{equation}

This design specifically addresses layer-wise heterogeneity 
through dynamic responsiveness: $\tau$ increases when 
current gradient norms exceed historical values 
(enhancing exploration for diverse features), 
while decreasing when gradients diminish 
(accelerating semantic-specific convergence). 
Refer to \cref{supp:train} for the detailed procedure 
applied during each training epoch. 
To continuesly maintain gradient propagation, 
following steps require matrix operations:

\begin{equation}\label{eq:5}
   \mathbb{M}'_{s,c} = \delta(c,\mathbf{S}_s)\quad \forall s \in [1, C_{(\text{s})}],  c \in [1, C]
\end{equation}
\begin{equation}\label{eq:6}
   \mathbb{M}_{s} = \mathbb{M'}_s\odot\mathbf{M^H}
   \quad s\in C_{(\text{s})}
\end{equation}

\noindent where $\mathbb{M}$ denotes a one-hot mapping matrix 
from input channels to selected channels, $\delta(a,b)$ denotes 
Kronecker delta function. Given \cref{eq:2} and 
\cref{eq:6}, we can finally obtain $\mathbf{Y}$ in \cref{eq:2} 
with gradient computation graph:

\begin{equation}
   \begin{array}{l}
      \mathbf{Y}=
      \mathbf{X}\oplus
      \{
         \mathbf{Z}_i\odot\mathbf{Z}_j
         \mid
         \{(i,j)\in\{1,2,\dots,C_{(\text{s})}\}, i\neq j\}
      \}
      \\[4pt]
      \text{s.t.}\quad
      \mathbf{Z}=\mathbb{M}\cdot\mathbf{X}
   \end{array}
\end{equation}

\noindent For inference stage, it directly takes the first 
few bits of the output of the ECA module and uses this 
as the index to extract the channels that need to be calculated, 
saving unnecessary calculation.

\subsection{DySoft Normalization}

The cross-Hadamard product creates input-adaptive channel 
combinations that enhance nonlinearity but produce unstable 
output distributions. Unlike conventional convolutions that 
rely on statistical normalization, this dynamic behavior 
renders batch normalization \citep{ioffe2015batch} ineffective 
and risks gradient explosion. Inspired by recent success of 
activation-based normalization in Transformers 
\citep{zhu2025cvpr}, we propose DySoft, a dynamic softsign 
normalization that intrinsically bounds outputs while 
maintaining hardware efficiency:

\begin{equation}
   y = \cfrac{\alpha x}{1+|\alpha x|}\cdot w+b
\end{equation}

\noindent where $\alpha,w,b$ denote learnable factors of an affine 
transform. Empirical comparisons \cref{tab:comparision-dyS} show 
softsign outperforms tanh and algebraic sigmoid variants in 
stability and computational efficiency, making it ideal for 
mobile deployment. The indispensability of DySoft is discussed in \cref{supp:dysoft}.

\begin{table}[!t]
   \begin{minipage}[!t]{0.49\textwidth}
      \centering
      \caption{
         \textbf{Comparison of dynamic sigmoidal curves.}
         The experimental conditions are the experimental 
         results of replacing the normalized layers of all 
         cross Hadamard products of the small model finally 
         determined in \cref{sec:exp}.
         }
      \scalebox{0.85}
      {
         \begin{tabular}{lccc}
         \toprule
         &
         \small Sigmoid & 
         \small Softsign & 
         \small Alge. Sigmoid \\
         \midrule
         Formula 
         & $\tfrac{e^x}{e^x + 1}$ 
         & $\tfrac{x}{1 + |x|}$ 
         & $\tfrac{x}{\sqrt{1 + x^2}}$ \\
         \midrule
         Top1(\%) & 73.14 & \textbf{73.57} & 72.80 \\
         \bottomrule
         \end{tabular}
      }
      \label{tab:comparision-dyS}
      \vspace{10pt}
      \caption{
         \textbf{Performance Comparison of ACH Module Replacement 
         on MobileNetV3.}
         There are a total of 11 Inverted Bottleneck modules in the 
         network, with indices starting from 0 in the table. 
         Several modules were selected for the ablation experiment. 
         The first row of the table represents the replaced layer(s), 
         and the second row represents the Top1 accuracy (\%). 
         `/' denotes the original unmodified MobileNetV3-S, `IB' denotes 
         Inverted Bottleneck. 
      }
      \scalebox{0.8}
      {
         \begin{tabular}{ccccccc}
         \toprule
         \small / & 
         \small \underline{IB\textsuperscript{0}} & 
         \small IB\textsuperscript{1} & 
         \small IB\textsuperscript{9} & 
         \small \underline{IB\textsuperscript{8}} & 
         \small IB\textsuperscript{10} & 
         \small IB\textsuperscript{9,10}\\
         \midrule
         \small 70.01 & \small 69.74 & \small 69.74 & \small 69.89 & 
         \small 70.38 & \small 71.03 & \small \textbf{71.58}\\
         \bottomrule
         \end{tabular}
      }
      \label{tab:placement}
   \end{minipage}
   \hfill
   \begin{minipage}[!t]{0.49\textwidth}
      \centering
      \caption{
         \textbf{Neural Architecture Search Confidence Distribution.}
         Showing selection confidence between Ghost and ACH variants. 
         Underlined channels indicate downsampling layers. Fixed layers (no search) marked with hyphens.
         Values below 0.01\% are indicated as \textless0.01\%. 
         DySig, DyAlge represent dynamic sigmoid and algebraic 
         sigmoid, as the abbreviation DySoft, respectively.
         }
      \scalebox{0.80}
      {
         \begin{tabular}{ccccc}
         \toprule
         \small Channels & 
         \small Ghost Conf. & 
         \multicolumn{3}{c}{\small ACH Variants Conf.} \\
         \cmidrule(lr){3-5}
         & & \small DySoft & \small DySig & \small DyAlge \\
         \midrule
         \underline{32} & - & - & - & - \\
         \underline{48} & - & - & - & - \\
         32 & - & - & - & - \\
         64 & \textbf{99.97\%} & \textless0.01\% & \textless0.01\% & \textless0.01\% \\
         64 & \textbf{99.99\%} & \textless0.01\% & \textless0.01\% & \textless0.01\% \\
         96 & \textbf{99.87\%} & \textless0.01\% & \textless0.01\% & \textless0.01\% \\
         96 & \textbf{99.69\%} & 0.24\% & \textless0.01\% & \textless0.01\% \\
         96 & \textbf{99.48\%} & 0.18\% & 0.05\% & 0.23\% \\
         96 & \textbf{98.40\%} & 0.80\% & 0.15\% & 0.51\% \\
         96 & \textbf{97.37\%} & 0.63\% & 0.63\% & 1.10\% \\
         96 & 2.92\% & \textbf{62.02\%} & 6.58\% & 28.18\% \\
         \underline{128} & 12.78\% & \textbf{43.72\%} & 23.87\% & 19.62\% \\
         128 & 0.34\% & \textbf{70.05\%} & 25.34\% & 4.24\% \\
         128 & 1.68\% & \textbf{34.56\%} & 29.40\% & 34.30\% \\
         960 & - & - & - & - \\
         \bottomrule
         \end{tabular}
      }
      \label{tab:nas-result}
   \end{minipage}
\end{table}

\subsection{Hadamard Adaptive Network}

To systematically validate the efficacy, implementability and 
architectural compatibility of the proposed ACH module, we construct 
Hadaptive-Net (Hadamard Adaptive Network), a network family that 
serves as a testbed for ACH module. We employ gradient-based Neural 
Architecture Search (NAS) \citep{dong2019searching} not to produce a single, 
static architecture, but as a principled methodology to discover 
the optimal integration of ACH modules within a modern, efficient 
backbone. This approach allows us to objectively evaluate ACH's 
performance and unearth general design principles for its 
deployment, mitigating the biases of manual heuristic design.

Our search is informed by a preliminary analysis revealing that 
ACH is depth-dependent, performing best in late-stage 
layers (\cref{tab:placement}). We thus designed a search space 
co-integrating ACH with GhostNet-style modules, enabling NAS to 
select the optimal operator per layer. The search 
results (\cref{tab:nas-result}) confirm our hypothesis: ACH is 
preferentially selected over Ghost modules in high-dimensional 
spaces. The finalized Hadaptive-Net architectures are derived 
from these discovered principles, with full specifications in 
the \cref{supp:conf}.

\section{Implementation}\label{sec:impl}

Prior to additional experimentation, we must ensure the 
cross-Hadamard product, as a novel operator, attains its 
theoretical efficiency on CPU/GPU and other hardware. 

\subsection{Computational Complexity Analysis}\label{subsec:complex}

The computational complexities for 
expanding channel dimension with a feature map of $f\times f$ 
from $m$ to $n$ dimensions 
with $k\times k$ convolution are analyzed as below.
For inverted bottleneck:
\begin{equation}\label{eq:ib}
   \mathcal{O}(mn\cdot f^2)_{\mathrm{pointwise\ conv}}
\end{equation}
For Ghost module, which partially replaces the expensive pointwise 
convolution with a more efficient strategy:
\begin{equation}\label{eq:ghost}
   \mathcal{O}(ms\cdot f^2)_{\mathrm{pointwise\ conv}}+
   \mathcal{O}((n-s)\cdot k^2f^2)_{\mathrm{cheap\ op}}
\end{equation}
Our method preserves the pointwise convolution while 
delegating channel expansion to Hadamard product operations:
\begin{equation}\label{eq:ach}
   \mathcal{O}(m^2\cdot f^2)_{\mathrm{pointwise\ conv}}+
   \mathcal{O}((n-m)\cdot f^2)_{\mathrm{hadamard}}
\end{equation}
Since $m \ll n$, 
The computational complexity of Ghost module 
is reduced to $\frac{s}{n}$ of inverted bottleneck convolution, 
while our ACH module achieves approximately $\frac{1}{m}$ of the 
inverted bottleneck convolution's complexity in channel expansion.
Derivations are shown in \cref{supp:impl}. 
Remarkably, each Hadamard-derived feature map requires only 
$f^2$ FLOPs, achieving superior efficiency compared to 
conventional approaches. 

The emphasis on FLOPs over latency is driven by the necessity to 
maintain cross-platform compatibility and approximate theoretical 
performance limits. While such a prioritization can be readily 
implemented and validated in serial processing architectures, 
heterogeneous computing systems present significant challenges 
that necessitate extensive optimization efforts.

\subsection{GPU Acceleration}\label{subsec:gpu}

While lower theoretical computational complexity typically 
suggests faster inference speed, the actual GPU execution 
involves intricate scheduling by the CPU. The sophisticated 
channel mapping process in ACH module often gets 
decomposed into multiple sub-operations by inference 
frameworks, manifesting as frequent CPU-GPU synchronization 
and repeated kernel launches. The triangular computation 
pattern of $C_n^2$ combinations for cross-Hadamard products 
necessitates specialized operator design, 
for which we propose two optimization approaches:

\begin{enumerate}
    \item \textbf{Direct-Indexing}: Each thread 
    block exclusively handles one Hadamard product. 
    The closed-form mapping from pairing index $p$ to 
    to matrix indices $(i,j)$ is:
    \[
        \left\{
            \begin{array}{l}
                i = \cfrac{1}{2}\cdot\lfloor(2n-1)-\sqrt{(2n-1)^2-8p}\rfloor\\
                j = i + 1 + p - \cfrac{i\cdot(2n-1-i)}{2}\\
            \end{array}
        \right.
    \]
    where $n$ denotes number of candidate channels.
    \item \textbf{Parity-Balanced}: Assign c thread blocks 
    (c: input channels), evenly distributing irregular 
    computations via iterative indexing \cref{alg:parity}, 
    then compute pairing indices with an inverse formula:
    \[
        p = \cfrac{i\cdot(2n - i - 1)}{2} + (j - i - 1)
    \]
\end{enumerate}

\begin{wraptable}{r}{0.55\textwidth}
   \centering
   \captionsetup{skip=2pt}
   \caption{
      Acceleration performance.
      }
   \scalebox{0.8}
   {
    \centering
    \begin{tabular}{lccc}
    \toprule
    &
    (Native) & 
    Direct-Indexing & 
    Parity-Balanced \\
    \midrule
    Latency\ (ms)
    & $12.40$ 
    & $7.21$ 
    & $7.13$ \\
    \bottomrule
    \end{tabular}
   }
    \label{tab:acceleration}
\end{wraptable}

Table \ref{tab:acceleration} demonstrates the acceleration 
effects of different optimization approaches on the 
Hadaptive-Net-L, which confirms the indispensability of 
optimization in the step of implementation. 
To approach extreme performance, we analyzed the performance of 
the two algorithms under different characteristic scales. 
Details of the experiments and Parity-Balanced algorithm 
are shown in \cref{supp:impl}. Resulting in 
practical inference scenarios, it's recommended to employ 
the parity-balanced approach for high-channel/small-HW 
tensors, while considering direct-indexing for spatial 
dimensions near 32× multiples. For performance-critical 
applications, custom compilation of tilling strategies 
matching factors of specific spatial dimensions may be warranted.

\section{Experiment}
\label{sec:exp}

This section demonstrates the applied scenarios of our proposed method. 
All experiments on ACH module are modified from the configuration 
of image classification experiment.
All latency benchmarks were conducted within the ONNX Runtime \citep{onnxruntime}
framework. To demonstrate the practical optimization potential of 
our proposed ACH module, we implemented it as a custom CUDA operator. 
We stress that this result is presented to showcase the high 
optimizability of the ACH operator. It should not be interpreted as 
a strict, head-to-head speed comparison against baseline models, 
which utilize the standard, framework-provided operators 
native to ONNX Runtime.

\begin{figure}[t]
    \centering
    \begin{subfigure}[!t]{0.49\textwidth}
      \includegraphics[width=1\textwidth]{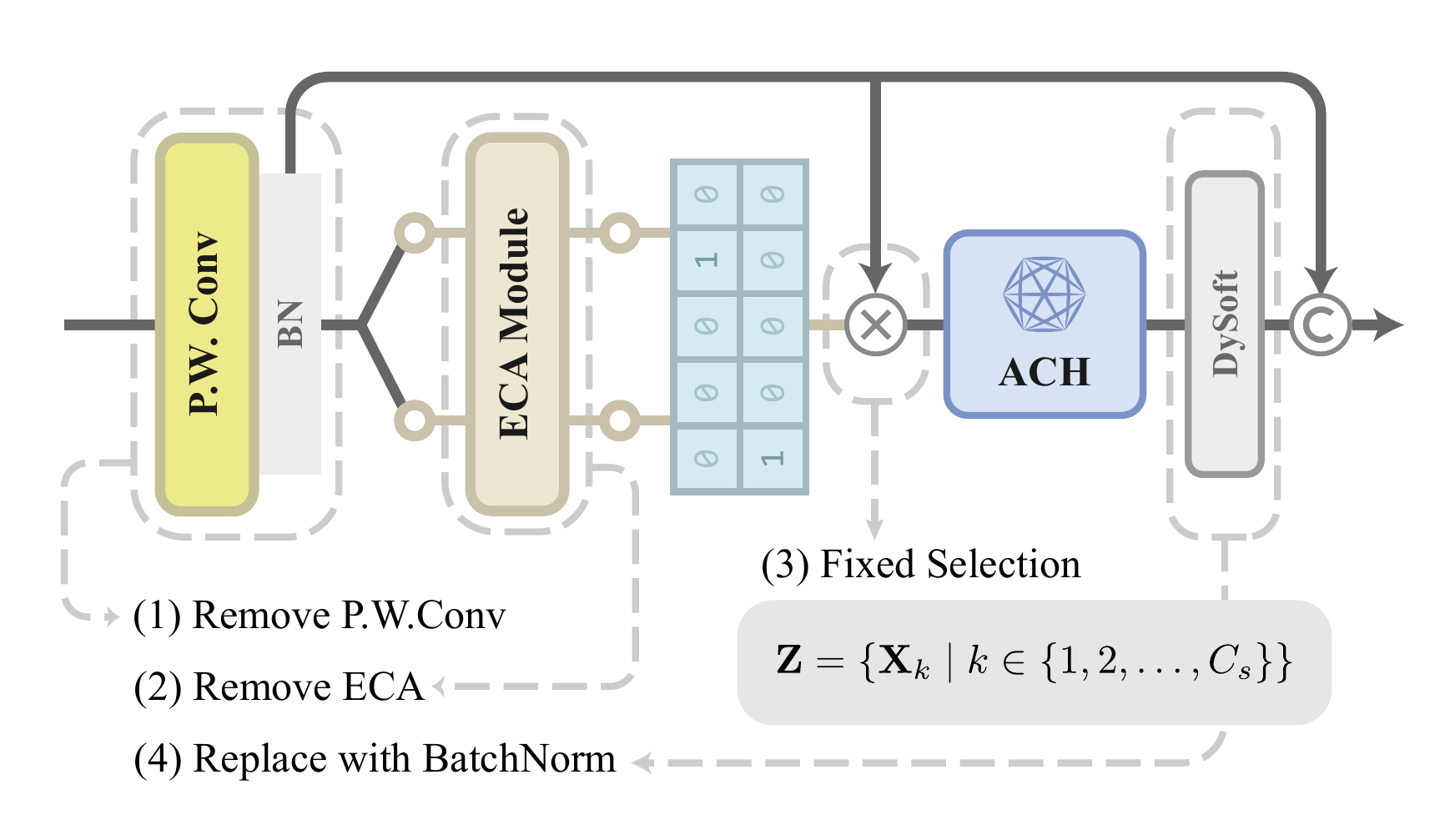}
    \end{subfigure}
    \hfill
    \begin{subfigure}[!t]{0.49\textwidth}
      \scalebox{0.9}
      {
      \begin{tabular}{ccccc}
      \toprule
      \small \textbf{P.W.Conv} & 
      \small \textbf{ECA} & 
      \small \textbf{Learnable} & 
      \small \textbf{DySoft} & 
      \small Top-1 \\
      \midrule
      \checkmark&&&\checkmark& \small69.27 \\
      \checkmark&&\checkmark&\checkmark& \small69.12 \\
      &\checkmark&\checkmark&\checkmark& \small71.96 \\
      \checkmark&\checkmark&\checkmark&& \small64.39 \\
      \checkmark&\checkmark&\checkmark&\checkmark& \small\textbf{73.57} \\
      \bottomrule
      \end{tabular}
      }
    \end{subfigure}
    \caption{
      \textbf{Component-wise ablation.}
      Illustration of component-wise ablation variations with 
      component-accuracy table. (1) and (2) represent removal of 
      pointwise convolution and ECA module, respectively. (3) 
      represents the replacement of learnable selection with fixed 
      channel combinations, and (4) represents the substitution of 
      cross-Hadamard normalization with standard batch normalization.
    }
   \label{tab:component-ablation}
\end{figure}

\subsection{Ablation on ACH Module}\label{subsec:ablation}

This experiment evaluates the contribution of each 
ACH component through controlled ablations: 
(1) Whether to keep pointwise convolution layer. 
(2) Whether to keep ECA module.
(3) Learnable selection or fixed combinations. 
(4) Dynamic softsign or batch normalization. 
The baseline model of this set of experiments is obtained from the 
best model of the previous set of experiments. 
Fig. \ref{tab:component-ablation} illustrates the variations of the ablation 
experiments with presenting the 
quantitative results, revealing that all the components 
serve their respective functions. 
The pointwise convolution provides fundamental channel-wise 
information exchange, while the ECA module enables the 
assessment of channel importance. These two components 
establish the essential foundation for the module's 
learnability. Disabling this learnability nearly renders 
the module ineffective, demonstrating that the discrete 
differentiation mechanism can properly provide gradients 
for the former components.The employment of dynamic 
softsign effectively circumvents gradient explosion risks, 
consequently exhibiting markedly better performance than 
batch normalization in experimental trials.

\begin{table}[!t]
   \begin{minipage}[t]{0.49\textwidth}
      \caption{
         \textbf{Replacements of ACH module on efficient models.}
         We replace the last two layers of each model. For instance, 
         replacing last two universal inverted bottleneck modules 
         for MobileNetV4.
      }
      \scalebox{0.8}{
      \centering
      \begin{tabular}{llll}
      \toprule
      Model & Top-1 & Params & FLOPs\\
      & (\%) & (M) &  (M)\\
      \midrule
      MobileNetV3-S & 70.01 & 1.61 & 123\\
      MobileNetV3-S (repl.) & 71.58$\uparrow$ & 1.55$\downarrow$& 114$\downarrow$\\
      \midrule
      MobileNetV4-S & 73.15 & 2.62 & 385\\
      MobileNetV4-S (repl.) & 72.19$\downarrow$ & 2.98$\uparrow$ & 381$\downarrow$\\
      \midrule
      ShuffleNetV2-1.0 & 65.89 & 1.36 & 303\\
      ShuffleNetV2-1.0 (repl.) & 71.68$\uparrow$ & 1.28$\downarrow$ & 291$\downarrow$\\
      \midrule
      StarNet-S1 & 71.84 & 2.68 & 854\\
      StarNet-S1 (repl.) & 72.07$\uparrow$ & 2.56$\downarrow$ & 810$\downarrow$\\
      \bottomrule
      \end{tabular}
      }
      \label{tab:replacement}
   \end{minipage}
   \hfill
   \begin{minipage}[t]{0.49\textwidth}
      \caption{
         \textbf{Performance of Hadaptive-Net on object detection.}
         We employ the SSD object detector to different scales of 
         Hadaptive-Net and baseline models with 
         COCO \citep{cocodataset} dataset.
      }
      \centering
      \begin{tabular}{lcc}
      \toprule
      \textbf{Backbone} & \textbf{mAP@0.5:0.95} & \textbf{mIOU} \\
      \midrule
      MobileNetV3-S & 21.7 & 71.2 \\
      MobileNetV2-1.0 & 21.9 & 70.5 \\
      GhostNetV3-1.0 & 22.7 & 72.8 \\
      \midrule
      Hadaptive-Net-S & 22.1 & 72.4 \\
      Hadaptive-Net-M & 22.9 & 73.0 \\
      Hadaptive-Net-L & \textbf{23.2} & \textbf{73.4} \\
      \bottomrule
      \end{tabular}
      \label{tab:coco}
   \end{minipage}
\end{table}


\begin{table*}[!t]
   \centering
   \caption{
      \textbf{Comparison of efficient models.}
      This table presents parameter counts, computational complexity 
      (FLOPs), and latency measurements obtained from the 
      CIFAR-100 \citep{Krizhevsky09learningmultiple}
      dataset.
      }
   \scalebox{0.8}
   {
   \begin{tabular}{l|cc|ccc|c|c}
   \toprule
   \multirow{3}*{\textbf{Model}} & & & 
   \multicolumn{3}{c}{\textbf{Latency}} & 
   \multicolumn{2}{c}{\textbf{Top-1 Accuracy}} \\
   ~  & \textbf{Params} & \textbf{FLOPs} & 
   \textbf{GPU} & \textbf{CPU} & \textbf{Mobile} & \textbf{CIFAR-100} & \textbf{ImageNet-1k}\\
   ~  & (M) & (M) & (ms) & (ms) & (ms) & (\%) & (\%) \\
   \midrule
   MobileNetV3-S \citep{howard2019searching} & 1.62 & 56 & 4.98 & 33.21 & 6.71 & 70.01 & 67.42 \\
   MobileOne-S0 \citep{AnasosaluVasu2022MobileOneAI} & 2.10 & 275 & 3.48 & 30.70 & 5.11 & 69.70 & 71.42 \\
   Hadaptive-Net-S (ours) & 2.10 & 131 & 3.41 & 29.45 & 4.28 & \textbf{73.57} & \textbf{73.96} \\
   ShuffleNetV2-1.0 \citep{ma2018shufflenet} & 2.28 & 146 & 3.58 & 40.60 & 4.55 & 65.89 & 69.40 \\
   MobileNetV4-S \citep{qin2024mobilenetv4} & 2.62 & 185 & 4.46 & 24.68 & 4.31 & 73.15 & 73.80 \\
   StarNet-S1 \citep{ma2024rewrite} & 2.68 & 422 & 6.00 & 82.73 & 7.96 & 71.84 & 73.50 \\
   \midrule
   Hadaptive-Net-M (ours) & 3.09 & 339 & 5.26 & 39.81 & 6.47 & \textbf{74.10} & \textbf{78.07} \\
   StarNet-S2 \citep{ma2024rewrite} & 3.43 & 544 & 7.30 & 94.50 & 8.41 & 67.70 & 74.78 \\
   GhostNet-1.0 \citep{han2020ghostnet} & 4.03 & 140 & 9.35 & 87.59 & 10.02 & 72.01 & 73.21 \\
   MobileNetV3-L \citep{howard2019searching} & 4.33 & 215 & 6.09 & 56.41 & 6.66 & 72.81 & 75.20 \\
   MobileOne-S1 \citep{AnasosaluVasu2022MobileOneAI} & 4.82 & 825 & 7.57 & 40.76 & 7.98 & 72.97 & 75.90 \\
   StarNet-S3 \citep{ma2024rewrite} & 5.49 & 754 & 8.60 & 112.7 & 9.87 & 68.27 & 77.26 \\
   \midrule
   Hadaptive-Net-L (ours) & 6.11 & 669 & 7.13 & 57.62 & 9.11 & \textbf{74.73} & 80.79 \\
   StarNet-S4 \citep{ma2024rewrite} & 7.22 & 1050 & 9.20 & 134.0 & 12.24 & 68.97 & 78.12 \\
   MobileOne-S2 \citep{AnasosaluVasu2022MobileOneAI} & 7.80 & 1299 & 9.77 & 61.19 & 10.23 & 73.25 & 77.41 \\
   GhostNetV3-1.0 \citep{Liu2024GhostNetV3ET} & 8.13 & 404 & 13.91 & 180.54 & 19.07 & 73.20 & 73.92 \\
   MobileNetV4-M \citep{qin2024mobilenetv4} & 8.56 & 827 & 8.42 & 47.93 & 9.36 & 74.66 & 79.88 \\
   MobileOne-S3 \citep{AnasosaluVasu2022MobileOneAI} & 10.15 & 1896 & 10.02 & 81.28 & 10.36 & 73.80 & 78.09 \\
   MobileNetV4-L \citep{qin2024mobilenetv4} & 31.44 & 2170 & 10.75 & 79.94 & 11.71 & 74.38 & \textbf{82.92} \\

   \bottomrule
   \end{tabular}
   }
   \label{tab:comparision}
\end{table*}

\subsection{Plug-and-Play Versatility of ACH Module}\label{subsec:plug}

The ACH module's distinct mechanism enhances semantic feature 
representation, making it ideal for standalone integration.
We validate this by replacing the final two layers of four 
state-of-the-art efficient networks: 
MobileNetV3 \citep{howard2019searching}, 
MobileNetV4 \citep{qin2024mobilenetv4}, 
ShuffleNetV2 \citep{ma2018shufflenet}, 
and StarNet \citep{ma2024rewrite}, 
with ACH. As \cref{tab:replacement} shows ACH improves accuracy 
in all networks except MobileNetV4 while reducing computational 
costs, confirming its generalizability as a plug-and-play 
performance enhancer.
See \cref{supp:extended} for deeper analysis.

\subsection{Image Classification}
\label{subsec:img-class}

We evaluate the performance of Hadaptive-Net on image classification 
(CIFAR-100 \citep{Krizhevsky09learningmultiple}, ImageNet-1K \citep{deng2009imagenet}), conducting comprehensive comparisons with 
other state-of-the-art efficient models. 
Our experiments use PyTorch with AdamW optimizer 
(lr$=$0.001, momentum$=$0.9, weight decay$=$1e-4) and 
CrossEntropyLoss. Training employs cosine annealing 
with $5\%$ linear warmup over 200 epochs (batch$=$64, 
224×224 inputs). 
We conducted the experiments both on 
CIFAR-100 \citep{Krizhevsky09learningmultiple} and 
ImageNet-1K \citep{deng2009imagenet}. 
Latency tests use ONNX-converted models (batch=1), 
500-run average (Hardware details in \cref{supp:exp}).

\noindent\textbf{Result}: According to \cref{tab:comparision}, Hadaptive-Net achieves 
superior accuracy in the first two groups while maintaining 
relatively low computational requirements. 
Although MobileNetV4 demonstrates the best performance in the 
largest parameter group, this comes at the cost of significantly 
higher computational overhead.

\subsection{Object Detection}
\label{subsec:object-detect}

To validate the generalization capability of HadaptiveNet as a 
backbone network across different downstream tasks, 
we conduct object detection experiments using the SSD \citep{2015arXivSSD}
framework. All models are trained on COCO train2017 
\citep{cocodataset} with a fixed input resolution of 320×320 
for 120 epochs, employing synchronized SGD optimization 
(momentum=0.9, weight decay=5e-4) 
and cosine learning rate decay initialized at 0.02. 
The detection head utilizes focal loss ($\gamma=$2.0) for 
classification and smooth L1 loss for bounding box regression. 
Evaluation follows the standard COCO protocol reporting 
mAP@[0.5:0.95] on val2017. For implementation details see 
\cref{supp:exp}.

\noindent\textbf{Result}:
As shown in \cref{tab:coco}, Hadaptive-Net continues 
the high-level performance of image classification tasks in 
the extended task of object detection. This proves that 
Hadaptive-Net has a more general feature extraction ability.

\noindent\textbf{Justification}:
Maximizing the efficacy of the ACH module likely requires an 
end-to-end co-design. We believe the more profound opportunity 
presented by this study lies in leveraging the principles of 
structured, lightweight cross-channel interaction embodied by 
ACH to redesign bottleneck components like Feature Pyramid 
Networks (FPNs), focusing on efficiently fusing multi-scale 
feature information. This represents a highly promising 
direction for breaking the efficiency bottleneck of current 
detectors.

\subsection{Generality on Transformer}

The Multi-Head Self-Attention (MHSA) mechanism in Transformer 
\citep{vaswani2017attention} focuses on the N-dimension, i.e., 
the relationships between tokens, while the Feed-Forward Network 
(FFN) operates on the C-dimension, integrating semantic 
information carried and aggregated within individual tokens. 
The FFN typically follows a classic inverted bottleneck structure, 
where the ACH module can effectively play a role in computational 
compression.

To maintain research consistency, focusing on computer vision 
tasks and lightweight design, we have decided to supplement our 
experiments with improvements on the MobileViT 
\citep{DBLP:conf/iclr/MehtaR22} model. 
Specifically, we replaced the FFN layers of the middle four 
Transformer encoders in MobileViTs with the ACH module.

\begin{wraptable}{r}{0.65\textwidth}
   \centering
   \captionsetup{skip=2pt}
   \caption{
      Replacements of ACH module on MobileViTs.
   }
   \scalebox{0.82}
   {
    \centering
    \begin{tabular}{lcccc}
      \toprule
      Model & Params(M) & GFLOPs & Top1-Acc & Top5-Acc \\
      \midrule
      MobileViT-small & 4.55 & 2.879 & 71.70 & 92.16 \\
      (Replaced) & 4.40 & 2.822 & 72.42 & 92.20 \\
      \midrule
      MobileViT-x-small & 1.80 & 1.559 & 69.98 & 91.43 \\
      (Replaced) & 1.74 & 1.537 & 70.48 & 91.71 \\
      \midrule
      MobileViT-xx-small & 0.88 & 0.588 & 67.62 & 90.19 \\
      (Replaced) & 0.85 & 0.579 & 67.42 & 90.32 \\
      \bottomrule
    \end{tabular}
   }
    \label{tab:mobilevit}
\end{wraptable}

\noindent\textbf{Result}:
The comparative results are shown in \cref{tab:mobilevit}, 
which demonstrate that replacing half of the FFN layers 
with the ACH module yielded significant improvements. Notably, 
this enhancement was achieved without increasing the number of 
parameters or computational complexity (FLOPs), leading to 
superior performance on the CIFAR-100 dataset compared to the 
baseline. This experiment substantiates that the ACH module 
exhibits a promising level of generalizability within deep 
learning, particularly for the role of a 
channel feature extractor. 
For natural language processing related attempts, 
please refer to \cref{supp:extended}.


\section{Conclusion}
\label{conclu}

This work systematically transforms Hadamard products from 
auxiliary operations into specific deep learning primitives, 
culminating in the development of the novel Adaptive Cross-Hadamard 
(ACH) module and its integration into Hadaptive-Net. Theoretical and 
empirical analyses show ACH's superiority over depthwise separable 
convolutions in computational efficiency and representational 
capacity. Lastly, this work establishes Hadamard-based 
operations as a valuable direction for efficient deep learning 
architectures, offers insights for integrating novel 
mathematical operations into neural network design.

\section*{Reproducibility Statement}

For theoretical verification, refer to \cref{supp:impl} for 
computational complexity analysis. For implementation 
reproducibility, \cref{sec:impl} discusses the whole 
principle of engineering acceleration algorithm kits. 
For training details, \cref{supp:conf} shows the training 
configuration of hyperparameters and hardware set. 
The code to implement the module and models in this paper has been 
open source.


\bibliography{ref}

@String(CVPR= {IEEE Conf. Comput. Vis. Pattern Recog.})

@String(ECCV= {Eur. Conf. Comput. Vis.})

@String(ICLR = {Int. Conf. Learn. Represent.})

@String(CVPR  = {CVPR})

@String(ECCV  = {ECCV})

@String(ICLR  = {ICLR})

@article{krizhevsky2012imagenet,
  title={Imagenet classification with deep convolutional neural networks},
  author={Krizhevsky, Alex and Sutskever, Ilya and Hinton, Geoffrey E},
  journal={Advances in neural information processing systems},
  volume={25},
  year={2012}
}

@inproceedings{he2016deep,
  title={Deep residual learning for image recognition},
  author={He, Kaiming and Zhang, Xiangyu and Ren, Shaoqing and Sun, Jian},
  booktitle={Proceedings of the IEEE conference on computer vision and pattern recognition},
  pages={770--778},
  year={2016}
}

@article{Ding2021RepVGGMV,
  title={RepVGG: Making VGG-style ConvNets Great Again},
  author={Xiaohan Ding and X. Zhang and Ningning Ma and Jungong Han and Guiguang Ding and Jian Sun},
  journal={2021 IEEE/CVF Conference on Computer Vision and Pattern Recognition (CVPR)},
  year={2021},
  pages={13728-13737},
  url={https://api.semanticscholar.org/CorpusID:231572790}
}

@article{Iandola2016SqueezeNetAA,
  title={SqueezeNet: AlexNet-level accuracy with 50x fewer parameters and <1MB model size},
  author={Forrest N. Iandola and Matthew W. Moskewicz and Khalid Ashraf and Song Han and William J. Dally and Kurt Keutzer},
  journal={ArXiv},
  year={2016},
  volume={abs/1602.07360},
  url={https://api.semanticscholar.org/CorpusID:14136028}
}

@article{howard2017mobilenets,
  title={Mobilenets: Efficient convolutional neural networks for mobile vision applications},
  author={Howard, Andrew G and Zhu, Menglong and Chen, Bo and Kalenichenko, Dmitry and Wang, Weijun and Weyand, Tobias and Andreetto, Marco and Adam, Hartwig},
  journal={arXiv preprint arXiv:1704.04861},
  year={2017}
}

@inproceedings{sandler2018mobilenetv2,
  title={Mobilenetv2: Inverted residuals and linear bottlenecks},
  author={Sandler, Mark and Howard, Andrew and Zhu, Menglong and Zhmoginov, Andrey and Chen, Liang-Chieh},
  booktitle={Proceedings of the IEEE conference on computer vision and pattern recognition},
  pages={4510--4520},
  year={2018}
}

@inproceedings{howard2019searching,
  title={Searching for mobilenetv3},
  author={Howard, Andrew and Sandler, Mark and Chu, Grace and Chen, Liang-Chieh and Chen, Bo and Tan, Mingxing and Wang, Weijun and Zhu, Yukun and Pang, Ruoming and Vasudevan, Vijay and others},
  booktitle={Proceedings of the IEEE/CVF international conference on computer vision},
  pages={1314--1324},
  year={2019}
}

@inproceedings{qin2024mobilenetv4,
  title={MobileNetV4: universal models for the mobile ecosystem},
  author={Qin, Danfeng and Leichner, Chas and Delakis, Manolis and Fornoni, Marco and Luo, Shixin and Yang, Fan and Wang, Weijun and Banbury, Colby and Ye, Chengxi and Akin, Berkin and others},
  booktitle={European Conference on Computer Vision},
  pages={78--96},
  year={2024},
  organization={Springer}
}

@article{AnasosaluVasu2022MobileOneAI,
  title={MobileOne: An Improved One millisecond Mobile Backbone},
  author={Pavan Kumar Anasosalu Vasu and James Gregory Gabriel and Jeff J. Zhu and Oncel Tuzel and Anurag Ranjan},
  journal={2023 IEEE/CVF Conference on Computer Vision and Pattern Recognition (CVPR)},
  year={2022},
  pages={7907-7917},
  url={https://api.semanticscholar.org/CorpusID:257805169}
}

@inproceedings{tan2019mnasnet,
  title={Mnasnet: Platform-aware neural architecture search for mobile},
  author={Tan, Mingxing and Chen, Bo and Pang, Ruoming and Vasudevan, Vijay and Sandler, Mark and Howard, Andrew and Le, Quoc V},
  booktitle={Proceedings of the IEEE/CVF conference on computer vision and pattern recognition},
  pages={2820--2828},
  year={2019}
}

@inproceedings{tan2019efficientnet,
  title={Efficientnet: Rethinking model scaling for convolutional neural networks},
  author={Tan, Mingxing and Le, Quoc},
  booktitle={International conference on machine learning},
  pages={6105--6114},
  year={2019},
  organization={PMLR}
}

@inproceedings{zhang2018shufflenet,
  title={Shufflenet: An extremely efficient convolutional neural network for mobile devices},
  author={Zhang, Xiangyu and Zhou, Xinyu and Lin, Mengxiao and Sun, Jian},
  booktitle={Proceedings of the IEEE conference on computer vision and pattern recognition},
  pages={6848--6856},
  year={2018}
}

@inproceedings{ma2018shufflenet,
  title={Shufflenet v2: Practical guidelines for efficient cnn architecture design},
  author={Ma, Ningning and Zhang, Xiangyu and Zheng, Hai-Tao and Sun, Jian},
  booktitle={Proceedings of the European conference on computer vision (ECCV)},
  pages={116--131},
  year={2018}
}

@inproceedings{han2020ghostnet,
  title={Ghostnet: More features from cheap operations},
  author={Han, Kai and Wang, Yunhe and Tian, Qi and Guo, Jianyuan and Xu, Chunjing and Xu, Chang},
  booktitle={Proceedings of the IEEE/CVF conference on computer vision and pattern recognition},
  pages={1580--1589},
  year={2020}
}

@article{Tang2022GhostNetV2EC,
  title={GhostNetV2: Enhance Cheap Operation with Long-Range Attention},
  author={Yehui Tang and Kai Han and Jianyuan Guo and Chang Xu and Chaoting Xu and Yunhe Wang},
  journal={ArXiv},
  year={2022},
  volume={abs/2211.12905},
  url={https://api.semanticscholar.org/CorpusID:253801665}
}

@article{Liu2024GhostNetV3ET,
  title={GhostNetV3: Exploring the Training Strategies for Compact Models},
  author={Zhenhua Liu and Zhiwei Hao and Kai Han and Yehui Tang and Yunhe Wang},
  journal={ArXiv},
  year={2024},
  volume={abs/2404.11202},
  url={https://api.semanticscholar.org/CorpusID:269187756}
}

@InProceedings{Chen_2023_CVPR,
    author    = {Chen, Jierun and Kao, Shiu-hong and He, Hao and Zhuo, Weipeng and Wen, Song and Lee, Chul-Ho and Chan, S.-H. Gary},
    title     = {Run, Don't Walk: Chasing Higher FLOPS for Faster Neural Networks},
    booktitle = {Proceedings of the IEEE/CVF Conference on Computer Vision and Pattern Recognition (CVPR)},
    month     = {June},
    year      = {2023},
    pages     = {12021-12031}
}

@inproceedings{liu2022convnet,
  title={A convnet for the 2020s},
  author={Liu, Zhuang and Mao, Hanzi and Wu, Chao-Yuan and Feichtenhofer, Christoph and Darrell, Trevor and Xie, Saining},
  booktitle={Proceedings of the IEEE/CVF conference on computer vision and pattern recognition},
  pages={11976--11986},
  year={2022}
}

@inproceedings{woo2023convnext,
  title={Convnext v2: Co-designing and scaling convnets with masked autoencoders},
  author={Woo, Sanghyun and Debnath, Shoubhik and Hu, Ronghang and Chen, Xinlei and Liu, Zhuang and Kweon, In So and Xie, Saining},
  booktitle={Proceedings of the IEEE/CVF conference on computer vision and pattern recognition},
  pages={16133--16142},
  year={2023}
}

@article{vaswani2017attention,
  title={Attention is all you need},
  author={Vaswani, Ashish and Shazeer, Noam and Parmar, Niki and Uszkoreit, Jakob and Jones, Llion and Gomez, Aidan N and Kaiser, {\L}ukasz and Polosukhin, Illia},
  journal={Advances in neural information processing systems},
  volume={30},
  year={2017}
}

@article{dosovitskiy2020image,
  title={An image is worth 16x16 words: Transformers for image recognition at scale},
  author={Dosovitskiy, Alexey and Beyer, Lucas and Kolesnikov, Alexander and Weissenborn, Dirk and Zhai, Xiaohua and Unterthiner, Thomas and Dehghani, Mostafa and Minderer, Matthias and Heigold, Georg and Gelly, Sylvain and others},
  journal={arXiv preprint arXiv:2010.11929},
  year={2020}
}

@InProceedings{Chen_2022_CVPR,
    author    = {Chen, Yinpeng and Dai, Xiyang and Chen, Dongdong and Liu, Mengchen and Dong, Xiaoyi and Yuan, Lu and Liu, Zicheng},
    title     = {Mobile-Former: Bridging MobileNet and Transformer},
    booktitle = {Proceedings of the IEEE/CVF Conference on Computer Vision and Pattern Recognition (CVPR)},
    month     = {June},
    year      = {2022},
    pages     = {5270-5279}
}

@inproceedings{Pan2022EdgeViTsCL,
  title={EdgeViTs: Competing Light-weight CNNs on Mobile Devices with Vision Transformers},
  author={Junting Pan and Adrian Bulat and Fuwen Tan and Xiatian Zhu and Lukasz Dudziak and Hongsheng Li and Georgios Tzimiropoulos and Brais Mart{\'i}nez},
  booktitle={European Conference on Computer Vision},
  year={2022},
  url={https://api.semanticscholar.org/CorpusID:248572100}
}

@article{wang2024hada,
  title={HAda: Hyper-Adaptive Parameter-Efficient Learning for Multi-View ConvNets},
  author={Wang, Shiye and Li, Changsheng and Yan, Zeyu and Liang, Wanjun and Yuan, Ye and Wang, Guoren},
  journal={IEEE Transactions on Image Processing},
  year={2024},
  publisher={IEEE}
}

@inproceedings{huanghira,
  title={HiRA: Parameter-Efficient Hadamard High-Rank Adaptation for Large Language Models},
  author={Huang, Qiushi and Ko, Tom and Zhuang, Zhan and Tang, Lilian and Zhang, Yu},
  booktitle={The Thirteenth International Conference on Learning Representations},
  year={2025}
}

@article{yang2022focal,
  title={Focal modulation networks},
  author={Yang, Jianwei and Li, Chunyuan and Dai, Xiyang and Gao, Jianfeng},
  journal={Advances in Neural Information Processing Systems},
  volume={35},
  pages={4203--4217},
  year={2022}
}

@article{rao2022hornet,
  title={Hornet: Efficient high-order spatial interactions with recursive gated convolutions},
  author={Rao, Yongming and Zhao, Wenliang and Tang, Yansong and Zhou, Jie and Lim, Ser Nam and Lu, Jiwen},
  journal={Advances in Neural Information Processing Systems},
  volume={35},
  pages={10353--10366},
  year={2022}
}

@article{hochreiter1997long,
  title={Long short-term memory},
  author={Hochreiter, Sepp and Schmidhuber, J{\"u}rgen},
  journal={Neural computation},
  volume={9},
  number={8},
  pages={1735--1780},
  year={1997},
  publisher={MIT press}
}

@inproceedings{ma2024rewrite,
  title={Rewrite the stars},
  author={Ma, Xu and Dai, Xiyang and Bai, Yue and Wang, Yizhou and Fu, Yun},
  booktitle={Proceedings of the IEEE/CVF Conference on Computer Vision and Pattern Recognition},
  pages={5694--5703},
  year={2024}
}

@inproceedings{chen2022simple,
  title={Simple baselines for image restoration},
  author={Chen, Liangyu and Chu, Xiaojie and Zhang, Xiangyu and Sun, Jian},
  booktitle={European conference on computer vision},
  pages={17--33},
  year={2022},
  organization={Springer}
}

@inproceedings{dong2019searching,
  title={Searching for a robust neural architecture in four gpu hours},
  author={Dong, Xuanyi and Yang, Yi},
  booktitle={Proceedings of the IEEE/CVF conference on computer vision and pattern recognition},
  pages={1761--1770},
  year={2019},
}

@article{gumbel1954statistical,
  title={Statistical theory of extreme valuse and some practical applications},
  author={Gumbel, Emil Julius},
  journal={Nat. Bur. Standards Appl. Math. Ser. 33},
  year={1954}
}

@article{bengio2013estimating,
  title={Estimating or propagating gradients through stochastic neurons},
  author={Bengio, Yoshua},
  journal={arXiv preprint arXiv:1305.2982},
  year={2013}
}

@TECHREPORT{Krizhevsky09learningmultiple,
  author={Alex Krizhevsky},
  title={Learning multiple layers of features from tiny images},
  institution={University of Toronto},
  year={2009}
}

@inproceedings{deng2009imagenet,
  title={Imagenet: A large-scale hierarchical image database},
  author={Deng, Jia and Dong, Wei and Socher, Richard and Li, Li-Jia and Li, Kai and Fei-Fei, Li},
  booktitle={2009 IEEE conference on computer vision and pattern recognition},
  pages={248--255},
  year={2009},
  organization={Ieee}
}

@misc{onnxruntime,
  title={ONNX Runtime},
  author={ONNX Runtime developers},
  year={2021},
  howpublished={\url{https://onnxruntime.ai/}},
  note={Version: 1.20.1}
}

@inproceedings{ioffe2015batch,
  title={Batch normalization: Accelerating deep network training by reducing internal covariate shift},
  author={Ioffe, Sergey and Szegedy, Christian},
  booktitle={International conference on machine learning},
  pages={448--456},
  year={2015},
  organization={pmlr}
}

@article{DBLP:journals/corr/BaKH16,
  author       = {Lei Jimmy Ba and
                  Jamie Ryan Kiros and
                  Geoffrey E. Hinton},
  title        = {Layer Normalization},
  journal      = {CoRR},
  volume       = {abs/1607.06450},
  year         = {2016},
  url          = {http://arxiv.org/abs/1607.06450},
  eprinttype    = {arXiv},
  eprint       = {1607.06450},
  bibsource    = {dblp computer science bibliography, https://dblp.org}
}

@InProceedings{zhu2025cvpr,
    author    = {Zhu, Jiachen and Chen, Xinlei and He, Kaiming and LeCun, Yann and Liu, Zhuang},
    title     = {Transformers without Normalization},
    booktitle = {Proceedings of the IEEE/CVF Conference on Computer Vision and Pattern Recognition (CVPR)},
    month     = {June},
    year      = {2025},
    pages     = {14901-14911}
}

@InProceedings{Wang_2020_CVPR,
author = {Wang, Qilong and Wu, Banggu and Zhu, Pengfei and Li, Peihua and Zuo, Wangmeng and Hu, Qinghua},
title = {ECA-Net: Efficient Channel Attention for Deep Convolutional Neural Networks},
booktitle = {Proceedings of the IEEE/CVF Conference on Computer Vision and Pattern Recognition (CVPR)},
month = {June},
year = {2020}
}

@ARTICLE{2015arXivSSD,
       author = {{Liu}, Wei and {Anguelov}, Dragomir and {Erhan}, Dumitru and {Szegedy}, Christian and {Reed}, Scott and {Fu}, Cheng-Yang and {Berg}, Alexander C.},
        title = "{SSD: Single Shot MultiBox Detector}",
      journal = {arXiv e-prints},
         year = 2015,
        month = dec,
          eid = {arXiv:1512.02325},
        pages = {arXiv:1512.02325},
          doi = {10.48550/arXiv.1512.02325},
archivePrefix = {arXiv},
       eprint = {1512.02325},
 primaryClass = {cs.CV},
       adsurl = {https://ui.adsabs.harvard.edu/abs/2015arXiv151202325L}
}

@article{cocodataset,
  author    = {Tsung{-}Yi Lin and Michael Maire and Serge J. Belongie and Lubomir D. Bourdev and Ross B. Girshick and James Hays and Pietro Perona and Deva Ramanan and Piotr Doll{'{a} }r and C. Lawrence Zitnick},
  title     = {Microsoft {COCO:} Common Objects in Context},
  journal   = {CoRR},
  volume    = {abs/1405.0312},
  year      = {2014},
  url       = {http://arxiv.org/abs/1405.0312},
  archivePrefix = {arXiv},
  eprint    = {1405.0312},
  bibsource = {dblp computer science bibliography, https://dblp.org}
}

@article{DBLP:journals/pami/ChrysosWPTC25,
  author       = {Grigorios G. Chrysos and
                  Yongtao Wu and
                  Razvan Pascanu and
                  Philip H. S. Torr and
                  Volkan Cevher},
  title        = {Hadamard Product in Deep Learning: Introduction, Advances and Challenges},
  journal      = {{IEEE} Trans. Pattern Anal. Mach. Intell.},
  volume       = {47},
  number       = {8},
  pages        = {6531--6549},
  year         = {2025},
  url          = {https://doi.org/10.1109/TPAMI.2025.3560423},
  doi          = {10.1109/TPAMI.2025.3560423},
  bibsource    = {dblp computer science bibliography, https://dblp.org}
}

@inproceedings{DBLP:conf/iclr/KimOLKHZ17,
  author       = {Jin{-}Hwa Kim and
                  Kyoung Woon On and
                  Woosang Lim and
                  Jeonghee Kim and
                  Jung{-}Woo Ha and
                  Byoung{-}Tak Zhang},
  title        = {Hadamard Product for Low-rank Bilinear Pooling},
  booktitle    = {5th International Conference on Learning Representations, {ICLR} 2017,
                  Toulon, France, April 24-26, 2017, Conference Track Proceedings},
  publisher    = {OpenReview.net},
  year         = {2017},
  url          = {https://openreview.net/forum?id=r1rhWnZkg},
  bibsource    = {dblp computer science bibliography, https://dblp.org}
}

@inproceedings{DBLP:conf/iclr/LiW00L00ZL24,
  author       = {Siyuan Li and
                  Zedong Wang and
                  Zicheng Liu and
                  Cheng Tan and
                  Haitao Lin and
                  Di Wu and
                  Zhiyuan Chen and
                  Jiangbin Zheng and
                  Stan Z. Li},
  title        = {MogaNet: Multi-order Gated Aggregation Network},
  booktitle    = {The Twelfth International Conference on Learning Representations,
                  {ICLR} 2024, Vienna, Austria, May 7-11, 2024},
  publisher    = {OpenReview.net},
  year         = {2024},
  url          = {https://openreview.net/forum?id=XhYWgjqCrV},
  bibsource    = {dblp computer science bibliography, https://dblp.org}
}

@article{DBLP:journals/corr/abs-2312-00752,
  author       = {Albert Gu and
                  Tri Dao},
  title        = {Mamba: Linear-Time Sequence Modeling with Selective State Spaces},
  journal      = {CoRR},
  volume       = {abs/2312.00752},
  year         = {2023},
  url          = {https://doi.org/10.48550/arXiv.2312.00752},
  doi          = {10.48550/ARXIV.2312.00752},
  eprinttype    = {arXiv},
  eprint       = {2312.00752},
  bibsource    = {dblp computer science bibliography, https://dblp.org}
}

@inproceedings{DBLP:conf/icml/ZhuL0W0W24,
  author       = {Lianghui Zhu and
                  Bencheng Liao and
                  Qian Zhang and
                  Xinlong Wang and
                  Wenyu Liu and
                  Xinggang Wang},
  title        = {Vision Mamba: Efficient Visual Representation Learning with Bidirectional
                  State Space Model},
  booktitle    = {Forty-first International Conference on Machine Learning, {ICML} 2024,
                  Vienna, Austria, July 21-27, 2024},
  publisher    = {OpenReview.net},
  year         = {2024},
  url          = {https://openreview.net/forum?id=YbHCqn4qF4},
  bibsource    = {dblp computer science bibliography, https://dblp.org}
}

@inproceedings{DBLP:conf/iclr/MehtaR22,
  author       = {Sachin Mehta and
                  Mohammad Rastegari},
  title        = {MobileViT: Light-weight, General-purpose, and Mobile-friendly Vision
                  Transformer},
  booktitle    = {The Tenth International Conference on Learning Representations, {ICLR}
                  2022, Virtual Event, April 25-29, 2022},
  publisher    = {OpenReview.net},
  year         = {2022},
  url          = {https://openreview.net/forum?id=vh-0sUt8HlG},
  bibsource    = {dblp computer science bibliography, https://dblp.org}
}

@inproceedings{DBLP:conf/naacl/DevlinCLT19,
  author       = {Jacob Devlin and
                  Ming{-}Wei Chang and
                  Kenton Lee and
                  Kristina Toutanova},
  editor       = {Jill Burstein and
                  Christy Doran and
                  Thamar Solorio},
  title        = {{BERT:} Pre-training of Deep Bidirectional Transformers for Language
                  Understanding},
  booktitle    = {Proceedings of the 2019 Conference of the North American Chapter of
                  the Association for Computational Linguistics: Human Language Technologies,
                  {NAACL-HLT} 2019, Minneapolis, MN, USA, June 2-7, 2019, Volume 1 (Long
                  and Short Papers)},
  pages        = {4171--4186},
  publisher    = {Association for Computational Linguistics},
  year         = {2019},
  url          = {https://doi.org/10.18653/v1/n19-1423},
  doi          = {10.18653/V1/N19-1423},
  bibsource    = {dblp computer science bibliography, https://dblp.org}
}

@inproceedings{DBLP:conf/emnlp/SocherPWCMNP13,
  author       = {Richard Socher and
                  Alex Perelygin and
                  Jean Wu and
                  Jason Chuang and
                  Christopher D. Manning and
                  Andrew Y. Ng and
                  Christopher Potts},
  title        = {Recursive Deep Models for Semantic Compositionality Over a Sentiment
                  Treebank},
  booktitle    = {Proceedings of the 2013 Conference on Empirical Methods in Natural
                  Language Processing, {EMNLP} 2013, 18-21 October 2013, Grand Hyatt
                  Seattle, Seattle, Washington, USA, {A} meeting of SIGDAT, a Special
                  Interest Group of the {ACL}},
  pages        = {1631--1642},
  publisher    = {{ACL}},
  year         = {2013},
  url          = {https://doi.org/10.18653/v1/d13-1170},
  doi          = {10.18653/V1/D13-1170},
  bibsource    = {dblp computer science bibliography, https://dblp.org}
}

@inproceedings{DBLP:conf/naacl/WilliamsNB18,
  author       = {Adina Williams and
                  Nikita Nangia and
                  Samuel R. Bowman},
  editor       = {Marilyn A. Walker and
                  Heng Ji and
                  Amanda Stent},
  title        = {A Broad-Coverage Challenge Corpus for Sentence Understanding through
                  Inference},
  booktitle    = {Proceedings of the 2018 Conference of the North American Chapter of
                  the Association for Computational Linguistics: Human Language Technologies,
                  {NAACL-HLT} 2018, New Orleans, Louisiana, USA, June 1-6, 2018, Volume
                  1 (Long Papers)},
  pages        = {1112--1122},
  publisher    = {Association for Computational Linguistics},
  year         = {2018},
  url          = {https://doi.org/10.18653/v1/n18-1101},
  doi          = {10.18653/V1/N18-1101},
  bibsource    = {dblp computer science bibliography, https://dblp.org}
}
\bibliographystyle{iclr2026_conference}

\appendix
\section{Appendix}

\subsection{Training Mechanism}\label{supp:train}

\noindent\textbf{$\tau$ Adjustment}:
We implement distinct temperature control mechanisms for 
ACH modules versus NAS due to fundamental differences in 
their training paradigms. For ACH modules distributed 
across network layers, which process heterogeneous 
features and semantics, we deliberately design a adaptive 
regulation algorithm based on gradient norm trends:
\vspace{5pt}
\begin{algorithm}[!htbp]
\caption{$\tau$ Adjustment via Gradient Norm Tracking}
\label{alg:tau_adj}
\textbf{Input}: Current gradient tensor $grad$, scaling factor $\alpha=0.01$ \\
\textbf{Parameter}: Historical gradient norm $\tau_{hist}$, current temperature $\tau$ \\
\textbf{Output}: Updated temperature $\tau$
\begin{algorithmic}[1]
\IF{$\tau_{hist} \neq 0 \land grad \neq \text{NULL}$}
    \STATE $\Delta \gets \begin{cases} 
        1 & \text{if } \|grad\|_2 \geq \tau_{hist} \\
        -1 & \text{otherwise}
    \end{cases}$
    \STATE $\tau_{new} \gets \tau \cdot (1 + \alpha \cdot \Delta)$
    \STATE $\tau \gets \text{CLAMP}(\tau_{new}, 0.01, 4.0)$
\ENDIF
\IF{$grad \neq \text{NULL}$}
    \STATE $\tau_{hist} \gets \|grad\|_2$
\ENDIF
\RETURN $\tau$
\end{algorithmic}
\end{algorithm}
\vspace{5pt}

\noindent For differentiable NAS (GDAS), which operates under 
fundamentally different optimization constraints, 
we retain the GDAS framework's global annealing strategy:

\begin{itemize}
	\item Training protocol: Architecture parameters 
	undergo periodic updates separate from model weights, 
	with gradient clipping ($\|\nabla\|\leq1.0$) ensuring stable convergence.
	\item Temperature scheduler: Implements predefined decay strategies:
	\[
	\tau_e = \begin{cases} 
	\tau_{\max} - (\tau_{\max} - \tau_{\min}) \cdot \frac{e}{E}\\
	\tau_{\max} \cdot \left( \frac{\tau_{\min}}{\tau_{\max}} \right)^{\frac{e}{E}}\\
	\tau_{\min} + 0.5(\tau_{\max} - \tau_{\min}) \cdot \left[1 + \cos\left(\pi \frac{e}{E}\right)\right]\\
	\end{cases}
	\]
\end{itemize}

\noindent The three formulas represent linear, exponential, cosine 
annealing, respectively. Consistent with GDAS methodology, 
$\tau$ remains within [0.1,4.0] throughout training.\\

\noindent\textbf{NAS Specifics}:
The neural architecture search process employs a 
dual-optimizer framework with distinct settings for model 
parameters and architecture parameters. Training executes 
over 250 epochs with a global batch size of 64, utilizing 
CUDA acceleration on a single GPU device (ID 0). 
The primary model optimizer is AdamW with base learning 
rate 0.003, momentum 0.9, and weight decay $1e-4$, coupled 
with CosineAnnealingLR scheduling for learning rate decay. 
Architecture parameters undergo separate optimization 
via AdamW with specialized learning rate $3e-4$ to 
accommodate their distinct gradient distributions.

\begin{minipage}[!t]{0.49\textwidth}
   \begin{figure}[H]
      \centering
      \includegraphics[width=\textwidth]{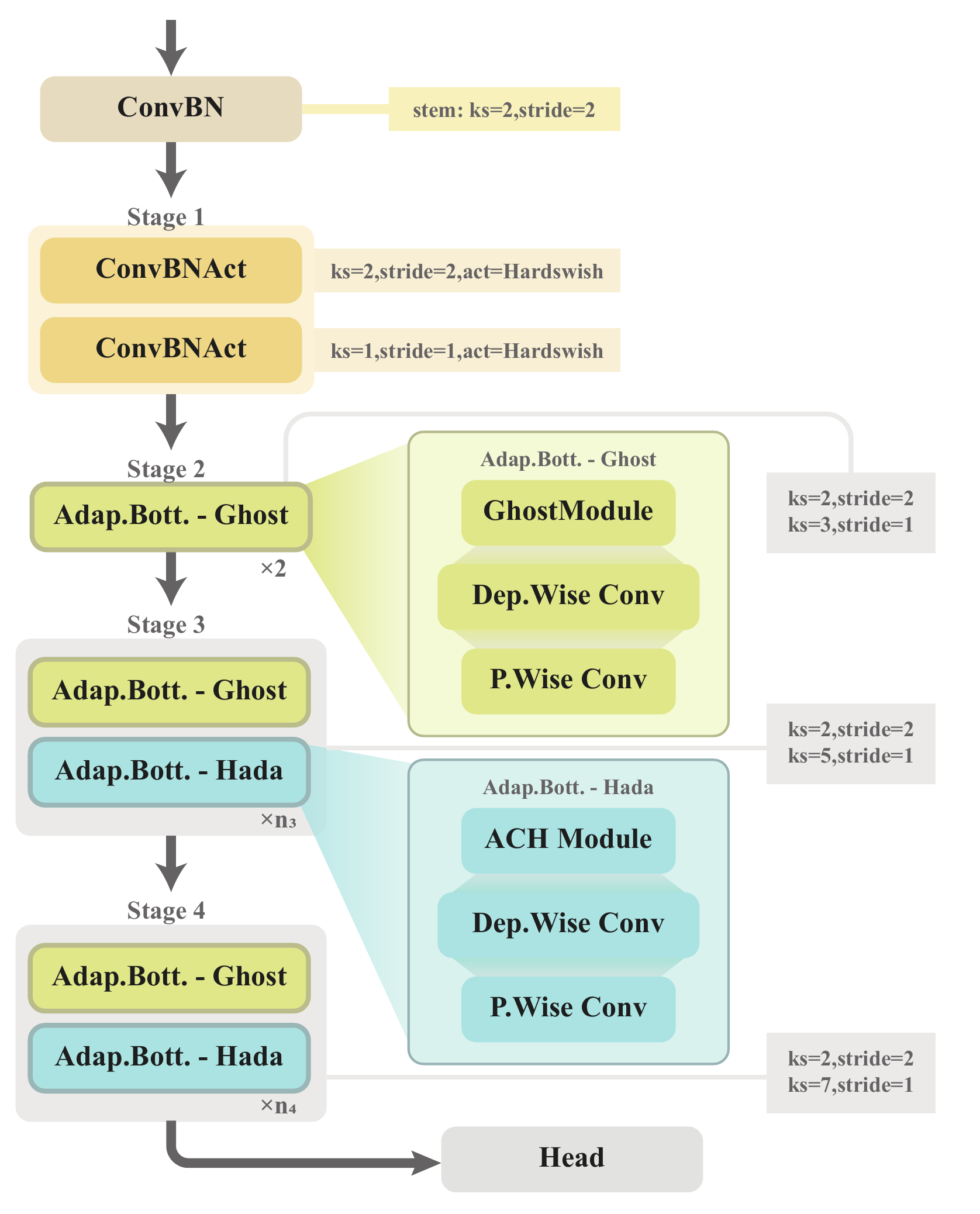}
      \caption{
         Hadaptive-Net architecture overview.
      }
      \label{fig:ov}
   \end{figure}
\end{minipage}
\hfill
\begin{minipage}[!t]{0.49\textwidth}
   \begin{table}[H]
   \centering
   \caption{
      Hadaptive-Net-S architecture details.
   }
   \label{tab:architecture}
   \scalebox{0.95}
   {
   \begin{tabular}{rrrll}
   \toprule
   Layer & Module & Arguments \\
   \midrule
   0 & CNA & [3, 32, 2, 2] \{BN, None\} \\
   1 & CNA & [32, 48, 2, 2] \{BN, HS\} \\
   2 & CNA & [48, 32, 1, 1] \{BN, HS\} \\
   3 & AB & [32, 64, 'Ghost', 4.0, 2, 2] \\
   4 & AB & [64, 64, 'Ghost', 2.0, 3, 1] \\
   5 & AB & [64, 96, 'Ghost', 4.0, 2, 2] \\
   6 & AB & [96, 96, 'Hada', 16, 5, 1] \\
   7 & AB & [96, 96, 'Hada', 16, 5, 1] \\
   8 & AB & [96, 96, 'Ghost', 2.0, 5, 1] \\
   9 & AB & [96, 96, 'Hada', 16, 5, 1] \\
   10 & AB & [96, 96, 'Hada', 16, 5, 1] \\
   11 & AB & [96, 128, 'Ghost', 6.0, 2, 2] \\
   12 & AB & [128, 128, 'Hada', 32, 7, 1] \\
   13 & AB & [128, 128, 'Hada', 32, 7, 1] \\
   14 & CNA & [128, 960, 1, 1] \{BN, HS\} \\
   15 & FN & [960, 100, 1280, 0.3] \\
   \bottomrule
   \end{tabular}
   }
   \end{table}
\end{minipage}

\subsection{DySoft Indispensability}\label{supp:dysoft}

Quantitative experiments can not explain the indispensability of 
DySoft since the model without DySoft training is extremely 
unstable and has no representative experimental data. However, 
these phenomena can illustrate a problem that DySoft is 
empirically necessary, which could be explained by 
probability theory.

\noindent\textbf{Problem Tracing}:
Traditional normalization methods, such as BatchNorm 
\citep{ioffe2015batch} and LayerNorm \citep{DBLP:journals/corr/BaKH16}, 
have a priori assumption that the statistical mean and 
statistical variance of the tensors they receive are knowable 
and traceable, which constitutes the basis of model convergence. 
In the process of ACH training and reasoning, we will involve a 
standard $\mathrm {Z}_i \odot\mathrm {Z}_j $ cross Hadamard 
product calculation. In previous machine learning methods, 
the use of Hadamard product is usually self referential, 
that is, $Z^2=\mathrm{Z}\odot\mathrm{Z}$. In this case, we can 
easily infer the mean value of $Z^2=\mathrm{Z}\odot\mathrm{Z}$ 
from the mean and variance $\mu,\sigma^2$ of $Z$ :

\[
\begin{array}{l}
\text{Var}(Z)=\text E[(Z-\mu)^2]=\text E[Z^2]-(\text E[Z])^2\\
E[Z^2]=\mu^2+\sigma^2
\end{array}
\]

Since tensor $Z$ was processed by normalization from above layer, 
which approximately satisfies $Z\sim N(\mu,\sigma^2)$. According 
to the fourth moment formula of normal distribution:

\[
\begin{array}{rcl}
\text E[Z^4]&=&\mu^4+6\mu^2\sigma^2+3\sigma^4\\
\text{Var}(Z^2)&=&\text E[Z^4]-(\text E[Z^2])^2\\
&=&(\mu^4+6\mu^2\sigma^2+3\sigma^4)-(\mu^2+\sigma^2)^2\\
&=&2\sigma^2(2\mu^2+\sigma^2)
\end{array}
\]

If the self referring Hadamard product is deformed, 
for example $\phi_1(Z)\odot \phi_2(Z)$, Let $\phi$ here be a 
linear transformation operator, the corresponding matrix form 
is $X_1,X_2\ (X\in\R^{m\times n})$, bias vectors 
are $b_1,b_2\ (b\in\R^{m})$, then:

\[
\text E[\phi(Z)]=\text E[XZ+b]=\mu\cdot\dfrac{\sum_i^m\sum_j^nX_{i,j}}{m}+\text E[b]
\]

For variance, since $Z$ can approximate normal distribution, 
here we assume that its elements are i.i.d, then there are:

\[
\text{Var}(\phi(Z))=\dfrac{1}{m}\cdot\sum_i^m\text{Var}(\phi(Z)_i)=\dfrac{1}{m}\cdot\sum_i^m\sum_j^nA_{i,j}^2\cdot\sigma^2=\sigma^2\cdot\dfrac{\|A\|^2_F}{m}
\]

Suppose $\phi$ is a nonlinear transformation operator, 
which does not directly exist the predictability of analytical 
solutions. However, the purpose of normalization method is not 
to accurately track the statistical representation of tensors, 
but to ensure that the statistical representation of tensors 
remains stable in the reasoning process.

Let the mapping $T_f:\R\times\R_{>0}\to\R\times\R_{\ge0}$ be: 
$(\mu, \sigma^2)\mapsto(\mu',\sigma'^2)$. If $T_f$ unbounded, 
that is, there is a sequence $(\mu_k,\sigma_k^2)$ such that 
$\|T_f(\mu_k,\sigma_k^2)\|\to\infty$ as long as a layer 
accidentally reaches the state (such as disturbance, 
initialization deviation), the next layer of statistics will be 
unstable; If $T_f$ is discontinuous or the derivative is 
unbounded (e.g. $f(z)=\mathrm 1_{z>0}$ is at $\mu=0$), small 
disturbance can lead to $\mu',\sigma'^2$ upheaval, resulting 
in unstable training.

BatchNorm is generally considered in CV tasks. 
BN independently estimates the mean and variance of $k$ for 
each channel:

\[
\hat\mu_k=\mathbb{E}_{\mathrm x\sim\mathcal B}[x_k],\quad\hat\sigma^2_k=\text{Var}_{\mathrm x\sim\mathcal B}(x_k)
\]

And perform channel by channel affine transformation:

\[
x_k'=\gamma_k\cdot\dfrac{x_k-\hat\mu_k}{\sqrt{\sigma^2_k+\epsilon}}+\beta_k
\]

This operation does not force statistical consistency between 
channels, but allows or even encourages significant statistical 
heterogeneity between channels:

\[
\exists\ i\neq j\quad\text{s.t.}\quad\hat\mu_i\neq\hat\mu_j,\ \hat\sigma^2_i\neq\hat\sigma^2_j
\]

This property is consistent with the inductive bias of 
"channel division" in convolutional networks - different channels 
can professionally respond to different semantic patterns 
(such as edge, texture, color), which is the key basis for its 
high representation efficiency in visual tasks. In contrast, 
LN is normalized in the sample dimension:

\[
\mathrm{x'}=\gamma\cdot\dfrac{\mathrm x-\mu}{\sigma}+\beta,\quad\mu=\dfrac{1}{C}\sum_k^Cx_k,\ \sigma^2=\dfrac{1}{C}\sum_k^C(x_k-\mu)^2
\]

The implicit priori is that all channels at the same spatial 
location should have the same statistical scale, which drives 
statistical convergence between channels. This assumption is 
compatible with the inductive bias of "all tokens are comparable" 
in the global attention mechanism (such as ViT), but in CNN 
dominated by local receptive fields, it will weaken the channel 
specific characterization ability and lead to performance 
degradation.

Let us consider $y_{ij}=x_i\odot x_j$, its output statistics 
depend on the joint second moment of the input channel. Under 
the heterogeneity distribution induced by BN, 
let $x_i\sim\mathcal N(\mu_i,\sigma^2_i),\ x_j\sim\mathcal N(\mu_j,\sigma^2_j)$ 
and i.i.d, then:

\[
\mathbb{E}[y_{ij}]=\mu_i\mu_j\\
\text{Var}(y_{ij})=\mu^2_i\sigma^2_j+\mu^2_j\sigma^2_i+\sigma^2_i\sigma^2_j
\]

When the channel statistics differ significantly 
(e.g. $|\mu_i|\gg|\mu_j|$ or $\sigma_i\gg\sigma_j$), the variance 
shows a multiplicative amplification effect, which is far beyond 
the single channel scale range. The affine parameters of BN are 
only channel specific, which can not effectively correct the new 
statistical offset caused by such cross-channel coupling. 
Otherwise, the pairing process of $i, j$ is obtained by the 
nonlinear transformation of each input, which makes it impossible 
for the statistical representation iterative map $T_f$ to find 
the fixed point.

Although LN normalization may be used inside ACH module, 
it is very important to understand the heterogeneity between 
channels in CV tasks. There is usually a typical CNN trunk 
containing BN upstream of the module, so the whole feature 
learning process has been dominated by the heterogeneity of BN 
a priori. The model's understanding of image semantics will 
evolve spontaneously towards the direction of 
"channel specialization". At this time, if a strong cross-channel 
nonlinear module with implicit homogeneity assumption is inserted 
into the reasoning chain, it will lead to a priori conflict.

\noindent\textbf{Solution}:
The DySoft we introduced is essentially a variant of the softsign 
activation function:

\[
y = \cfrac{\alpha x}{1+|\alpha x|}\cdot w+b,\quad\lim_{\alpha x\to\pm\infty}\cfrac{\alpha x}{1+|\alpha x|}=\pm1
\]

Due to the boundedness of softsign, no matter how large the input 
variance $\sigma^2$ is, the output variance is rigidly limited in 
the $(0,1)$ range; When the input is small, it shows approximate 
linearity and maintains the characteristics of the signal. 
The parameter $\alpha$ can dynamically balance the expression and 
compression of the layer.

When the cross-Hadamard $y_{ij}=x_i\odot x_j$ has variance 
like $\sigma_i,\sigma_j\gg1$, $\text{Var}(y_{ij})$ increased 
by $\mathcal O(\sigma^4)$. After accessing DySoft, this trend 
can be significantly compressed and given boundedness. 
In addition, DySoft is also designed based on the hypothesis of 
channel heterogeneity, which is a priori compatible with the 
heterogeneity of BN. Its $w, b$ parameters are channel specific, 
and can independently learn the scale and offset for each 
cross-Hadamard product channel. At the same time, it does not 
destroy the channel professional representation established by 
the upstream BN, and only makes local intervention on the 
"danger signal", thus realizing the organic unity of 
characterization and stability.

In summary, DySoft is a learnable statistical compression gating 
(SCG) module, which achieves hard variance clamping for 
high square error input through bounded nonlinear mapping 
$\mathcal S(u)=u/(1+|u|)$ and restores the characterization 
capacity in combination with channel 
specific affine transformation. Without violating the 
heterogeneity prior of batch normalization, the design 
effectively inhibits the growth of multiplicative variance 
caused by cross channel nonlinearity 
(such as cross Hadamard product), and makes the statistical 
map $T:(\mu, \sigma^2)\mapsto(\mu',\sigma'^2)$ bounded and 
smooth, so as to meet the core condition of "knowability", 
providing a stable and convergent statistical target for the 
normalization layer.

\subsection{Hadaptive-Net Configuration}\label{supp:conf}

This section mainly shows the results of three 
groups of NAS experiments and the decision of final 
Hadaptive-Net structure. See 
\cref{tab:nas-a,tab:nas-b,tab:nas-c} for NAS experiments details.

\begin{table}[t]
   \caption{
      \textbf{Neural Architecture Search Result (a).}
      Compared with different kernel sizes. Reaching 
      67.55\% top1-acc as result.
   }\label{tab:nas-a}
   \scalebox{1}
   {
      \tiny
      \begin{tabularx}{\linewidth}
         {
            >{\centering\arraybackslash}X>{\centering\arraybackslash}X
            >{\centering\arraybackslash}X>{\centering\arraybackslash}X
            >{\centering\arraybackslash}X>{\centering\arraybackslash}X
            >{\centering\arraybackslash}X>{\centering\arraybackslash}X
         }
      \toprule
      \small Channels & 
      \multicolumn{2}{c}{\small Ghost Conf.} & 
      \multicolumn{3}{c}{\small ACH Conf.} &
      \small Blank \\
      \cmidrule(lr){2-3} \cmidrule(lr){4-6}
      & \small 2 & \small 3 & \small 3 & \small 5 & \small 7 & \\
      \midrule
      \small \underline{32} & \small - & \small - & \small - & \small - \\
      \small \underline{48} & \small - & \small - & \small - & \small - \\
      \small 32 & \small - & \small - & \small - & \small - \\
      \small \underline{64} & \small 54\% & \small 46\% & \small - & \small - & \small - & \small - \\
      \small 64 & \small - & \small 24\% & \small - & \small 76\% & \small - & \small - \\
      \small \underline{96} & \small 52\% & \small 48\% & \small - & \small - & \small - & \small - \\
      \small 96 & \small 27\% & \small 14\% & \small 9\% & \small 50\% & \small - & \small - \\
      \small 96 & \small 25\% & \small 21\% & \small 28\% & \small 26\% & \small - & \small - \\
      \small 96 & \small 20\% & \small 17\% & \small 23\% & \small 17\% & \small - & \small 23\% \\
      \small 96 & \small 19\% & \small 19\% & \small 23\% & \small 17\% & \small - & \small 23\% \\
      \small 96 & \small 20\% & \small 20\% & \small 20\% & \small 18\% & \small - & \small 22\% \\
      \small \underline{128} & \small 37\% & \small 28\% & \small 17\% & \small 17\% & \small - & \small - \\
      \small 128 & \small - & \small - & \small 35\% & \small 33\% & \small 33\% & \small - \\
      \small 128 & \small - & \small - & \small 60\% & \small 27\% & \small 13\% & \small - \\
      \small 128 & \small - & \small - & \small 1\% & \small 1\% & \small 1\% & \small 96\% \\
      \small 128 & \small - & \small - & \small 1\% & \small 1\% & \small 1\% & \small 96\% \\
      \small 128 & \small - & \small - & \small 1\% & \small 2\% & \small 1\% & \small 96\% \\
      \small 960 & \small - & \small - & \small -   & \small -   & \small -   & \small - \\
      \bottomrule
      \end{tabularx}
   }
\end{table}

\begin{table}[t]
   \begin{minipage}[t]{0.49\textwidth}
   \centering
   \caption{
      \textbf{Neural Architecture Search Result (c).}
      Shows the distribution of ACH configurations across different channel sizes.
      Values represent percentage confidence (rounded to nearest integer).
      '-' indicates layers that were not searched.
      Format: ACH-[chosen\_dim]-[kernel\_size]. 
   Reaching 67.73\% top1-acc as result.
   }\label{tab:nas-c}
   \begin{tabular}{cccccc}
   \toprule
   \small Channels & 
   \multicolumn{4}{c}{\small ACH Conf.} \\
   \cmidrule(lr){2-5}
   & \small 16-3 & \small 16-5 & \small 32-3 & \small 48-3 \\
   \midrule
   \small \underline{32} & \small - & \small - & \small - & \small - \\
   \small \underline{48} & \small - & \small - & \small - & \small - \\
   \small 32 & \small - & \small - & \small - & \small - \\
   \small \underline{64} & \small - & \small - & \small - & \small - \\
   \small 64 & \small - & \small - & \small - & \small - \\
   \small \underline{96} & \small - & \small - & \small - & \small - \\
   \small 96 & \small 68\% & \small 12\% & \small 20\% & \small - \\
   \small 96 & \small 44\% & \small 13\% & \small 43\% & \small - \\
   \small 96 & \small 47\% & \small 16\% & \small 37\% & \small - \\
   \small 96 & \small 35\% & \small 20\% & \small 45\% & \small - \\
   \small 96 & \small 45\% & \small 15\% & \small 39\% & \small - \\
   \small \underline{128} & \small - & \small - & \small - & \small - \\
   \small 128 & \small 19\% & \small - & \small 52\% & \small 29\% \\
   \small 128 & \small 28\% & \small - & \small 32\% & \small 40\% \\
   \small 128 & \small 36\% & \small - & \small 41\% & \small 22\% \\
   \small 128 & \small 40\% & \small - & \small 30\% & \small 30\% \\
   \small 128 & \small 51\% & \small - & \small 23\% & \small 26\% \\
   \small 960 & \small - & \small - & \small - & \small - \\
   \bottomrule
   \end{tabular}
\end{minipage}
\hfill
\begin{minipage}[t]{0.49\textwidth}
   \centering
   \caption{
      \textbf{Neural Architecture Search Result (b).}
      Shows the distribution of ACH configurations across different channel sizes.
      Values represent percentage confidence (rounded to nearest integer).
      '-' indicates layers that were not searched. Since 
      training ACH modules requires a lot of iterations 
      to be effective, the network tends to skip 
      them during training. Reaching 66.23\% top1-acc as 
      result.
   }\label{tab:nas-b}
   \scalebox{1}
   {
      \begin{tabular}{cccccc}
      \toprule
      \small Channels & 
      \multicolumn{3}{c}{\small ACH Conf.} &
      \small Blank \\
      \cmidrule(lr){2-4}
      & \small 16 & \small 32 & \small 48 & \\
      \midrule
      \small \underline{32} & \small - & \small - & \small - & \small - \\
      \small \underline{48} & \small - & \small - & \small - & \small - \\
      \small 32 & \small - & \small - & \small - & \small - \\
      \small \underline{64} & \small - & \small - & \small - & \small - \\
      \small 64 & \small - & \small - & \small - & \small - \\
      \small \underline{96} & \small - & \small - & \small - & \small - \\
      \small 96 & \small 14\% & \small 9\% & \small - & \small 68\% \\
      \small 96 & \small 14\% & \small 13\% & \small - & \small 65\% \\
      \small 96 & \small 16\% & \small 13\% & \small - & \small 62\% \\
      \small 96 & \small 13\% & \small 15\% & \small - & \small 61\% \\
      \small 96 & \small 13\% & \small 14\% & \small - & \small 60\% \\
      \small \underline{128} & \small - & \small - & \small - & \small - \\
      \small 128 & \small 1\% & \small 1\% & \small 1\% & \small 98\% \\
      \small 128 & \small 1\% & \small 1\% & \small 1\% & \small 98\% \\
      \small 128 & \small 1\% & \small 1\% & \small 1\% & \small 98\% \\
      \small 128 & \small 1\% & \small 1\% & \small 1\% & \small 98\% \\
      \small 128 & \small 1\% & \small 1\% & \small 1\% & \small 98\% \\
      \small 960 & \small - & \small - & \small - & \small - \\
      \bottomrule
      \end{tabular}
   }
\end{minipage}
\end{table}

Hadaptive-Net adopts a hierarchical backbone architecture 
comprising a stem followed by four distinct stages, as 
shown in \cref{fig:ov}. 
To implement Ghost and ACH module with adaptability, 
we design the Adaptive Bottleneck that can decide the 
expansion layer of the bottleneck manually. 
The network begins with a linear convolutional layer as the 
stem, followed by fixed two conventional convolutional layers 
in Stage 1 for initial feature extraction. Stage 2 incorporates 
two fixed Adaptive Bottlenecks utilizing Ghost module as 
expansion layers, enabling rapid downsampling. 
Stages 3 and 4 employ Ghost Ada.Bott. for downsampling layers 
and Hadamard Ada.Bott for repeated residual blocks, 
with particular emphasis on parameter concentration in Stage 3, 
following ConvNeXt's design philosophy. The kernel sizes 
progressively increase across stages, with non-downsampling 
layers configured as $1\times1$, $3\times3$, 
$5\times5$, and $7\times7$ respectively. 

Refer to \cref{tab:architecture} for detailed description 
of layer level architecture configuration. 
CNA denotes combinition of convolution, normalization and
activation layers. AB denotes adaptive bottleneck, which 
could be subdivided into Ghost module or ACH 
(Adaptive Cross-Hadamard) module. Hada denotes the 
ACH module. BN denotes batch normalization. HS denotes 
hardswish activation. FN denotes full connection layer.
All first two arguments represent input/output channel. 
All last two arguments represents kernel size and stride 
size, respectively.

\subsection{Experiments Details}\label{supp:exp}

The following is a detailed description 
of the experimental part of this paper.\\

\noindent\textbf{Hardware Configurations}:
Latency tests conducted on:
\begin{itemize}  
    \item \textbf{Desktop GPU}: NVIDIA RTX TITAN (24GB GDDR6, CUDA 11.6)  
    \item \textbf{Server CPU}: Intel Xeon Gold 5218 (2.3GHz, 16C/32T)  
    \item \textbf{Mobile SoC}: Qualcomm Snapdragon 870 (4×Cortex-A77@2.4GHz + 1×A77@3.2GHz, Adreno 650)  
\end{itemize}  
\noindent All tests used ONNX Runtime 1.16.0 with default execution providers.\\

\noindent\textbf{Object Detection - Training Protocol}:
The base learning rate of 0.02 corresponds to a batch size 
of 64 distributed across 5 GPUs, scaled linearly according 
to the batch size. We apply 3-epoch linear warmup and 
reduce the learning rate to 1e-5 via cosine scheduling. 
Data augmentation includes random HSV color jittering with 
hue delta limited to 18 degrees and saturation scaling 
between 0.5-1.5, followed by random canvas expansion up 
to 2x original size and IoU-based cropping with thresholds 
sampled from [0.1,0.3,0.5].\\

\noindent\textbf{Object Detection - Architecture Specifications}
The SSD detector generates 6 default boxes per feature map 
location with aspect ratios spanning 
[1:1, 1:2, 1:3, 2:1, 3:1]. Feature maps are extracted from 
five backbone stages with strides of [8,16,32,64,128] 
pixels respectively, corresponding to spatial dimensions 
from 38×38 down to 1×1. During focal loss computation we 
set the $\alpha$-balancing parameter to 0.25 after empirical 
validation across the range [0.1,0.5].

\subsection{Implementation Details}\label{supp:impl}

This section will supplement the derivation of previous 
computational complexity analysis and implementation details of GPU 
acceleration algorithm mentioned in the original text. 

With \cref{eq:ib,eq:ghost,eq:ach}, we can derive the ratio of 
Ghost Module complexity to that of the standard 
pointwise convolution:

\begin{equation}
   \text{Ratio}_\text{Ghost}
   =\frac{m\cdot s\cdot f^2+(n-s)\cdot k^2\cdot f^2}{m\cdot n\cdot f^2}
   =\frac{s}{n}+\frac{n-s}{n}+\frac{k^2}{m}
\end{equation}

Since $s$ is often chosen as a fraction of $n$ 
(e.g., $s=n/2$), the term $\frac{n-s}{n}$ is approximately a 
constant (e.g., $1/2$). Since $k^2$ is small (e.g., 
$9$ for a $3\times3$ kernel) and m can be relatively large, 
the term $\frac{k^2}{m}$ is often negligible. 
Under the condition $m\ll n$, the simplified complexity ratio is:

\begin{equation}
   \text{Ratio}_\text{Ghost}\approx\frac{s}{n}
\end{equation}

Similarly, we calculate the efficiency ratio by comparing ACH module 
complexity to the standard pointwise convolution:

\begin{equation}
   \text{Ratio}_\text{ACH}
   =\frac{m^2\cdot f^2+(n-m)\cdot f^2}{m\cdot n\cdot f^2}
   =\frac{m^2+n-m}{m\cdot n}
   =\frac{m}{n}+\frac{1}{m}+\frac{1}{n}
\end{equation}

Given the constraint $m<n,1\ll m,1\ll n$, 
the terms $\frac{1}{m}$ and 
$\frac{1}{n}$ become very small and can be considered negligible. 
Thus, the simplified complexity ratio for the ACH module is:

\begin{equation}
   \text{Ratio}_\text{ACH}\approx\frac{m}{n}
\end{equation}

For this theoretical analysis, we have carried out several groups 
of measured data on different input channel sizes and 
amplification ratios to confirm that ACH module has a very 
strong reasoning speed compared with the standard point-by-point 
convolution and ghost module.

\begin{figure}[!t]
\begin{subfigure}[t]{0.49\textwidth}
  \centering
  \includegraphics[width=\textwidth]{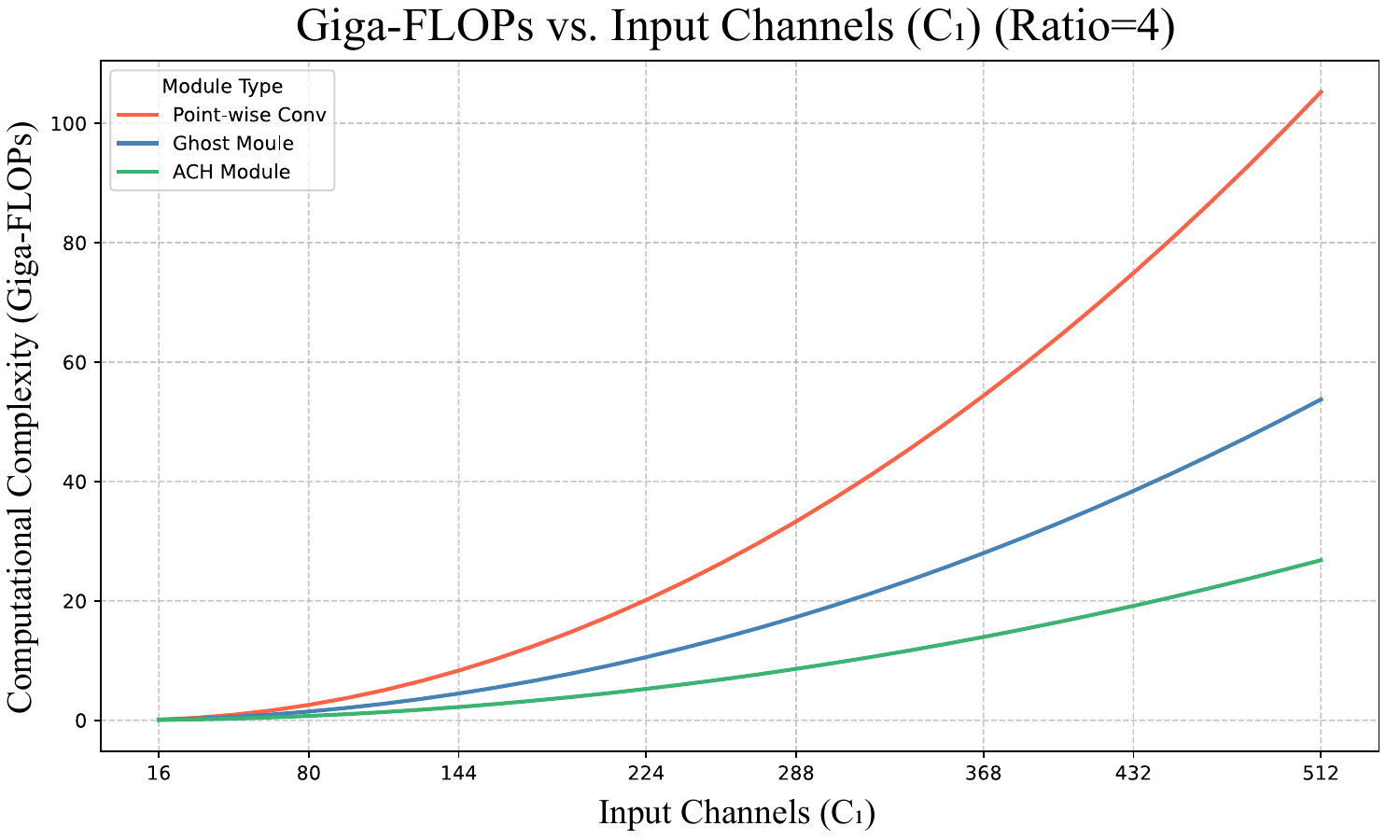}
\end{subfigure}
\hfill
\begin{subfigure}[t]{0.49\textwidth}
  \centering
  \includegraphics[width=\textwidth]{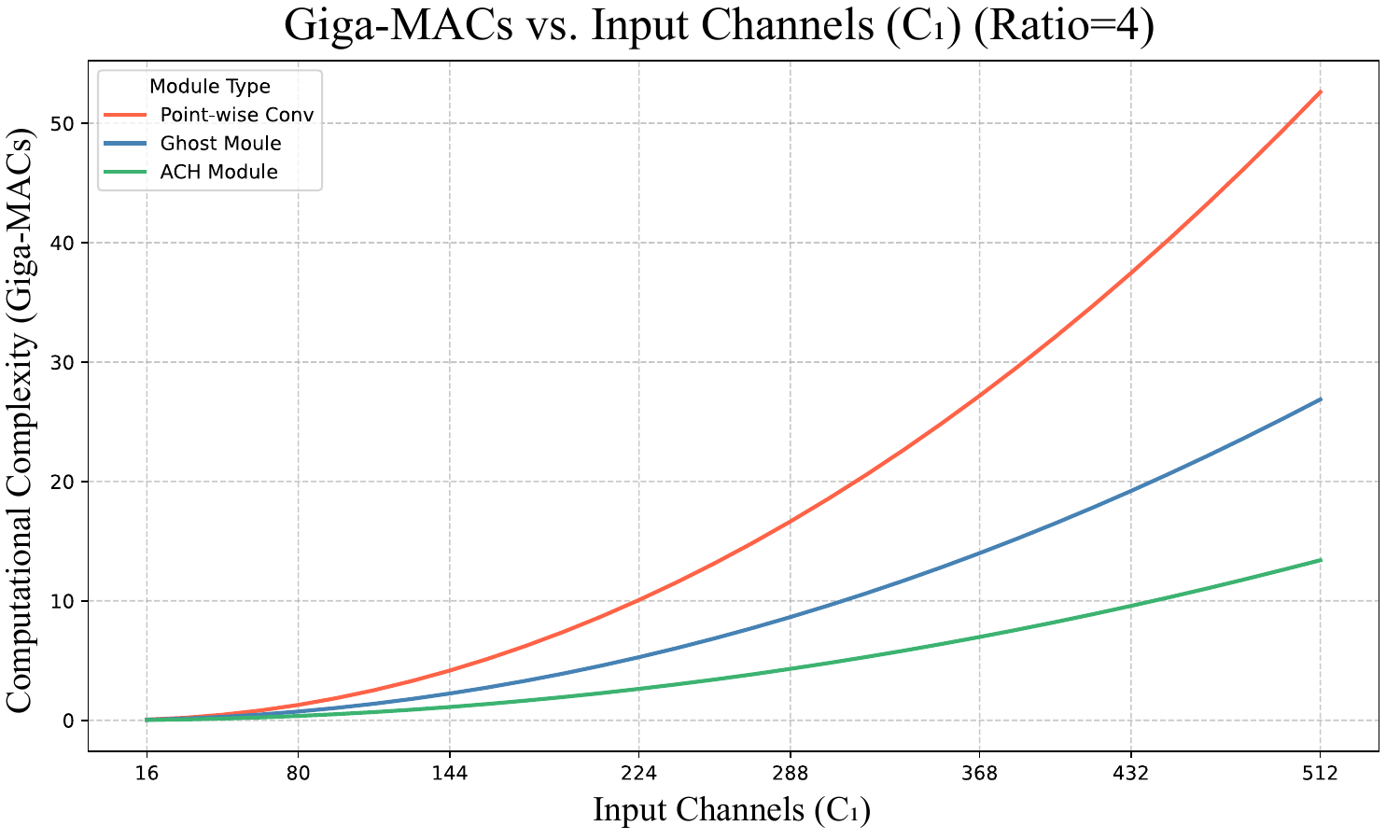}
\end{subfigure}
\caption{
   Comparison of computational efficiency under different input channel sizes
}
\label{fig:channel_calc}
\end{figure}

\begin{figure}[!t]
\begin{subfigure}[t]{0.49\textwidth}
  \centering
  \includegraphics[width=\textwidth]{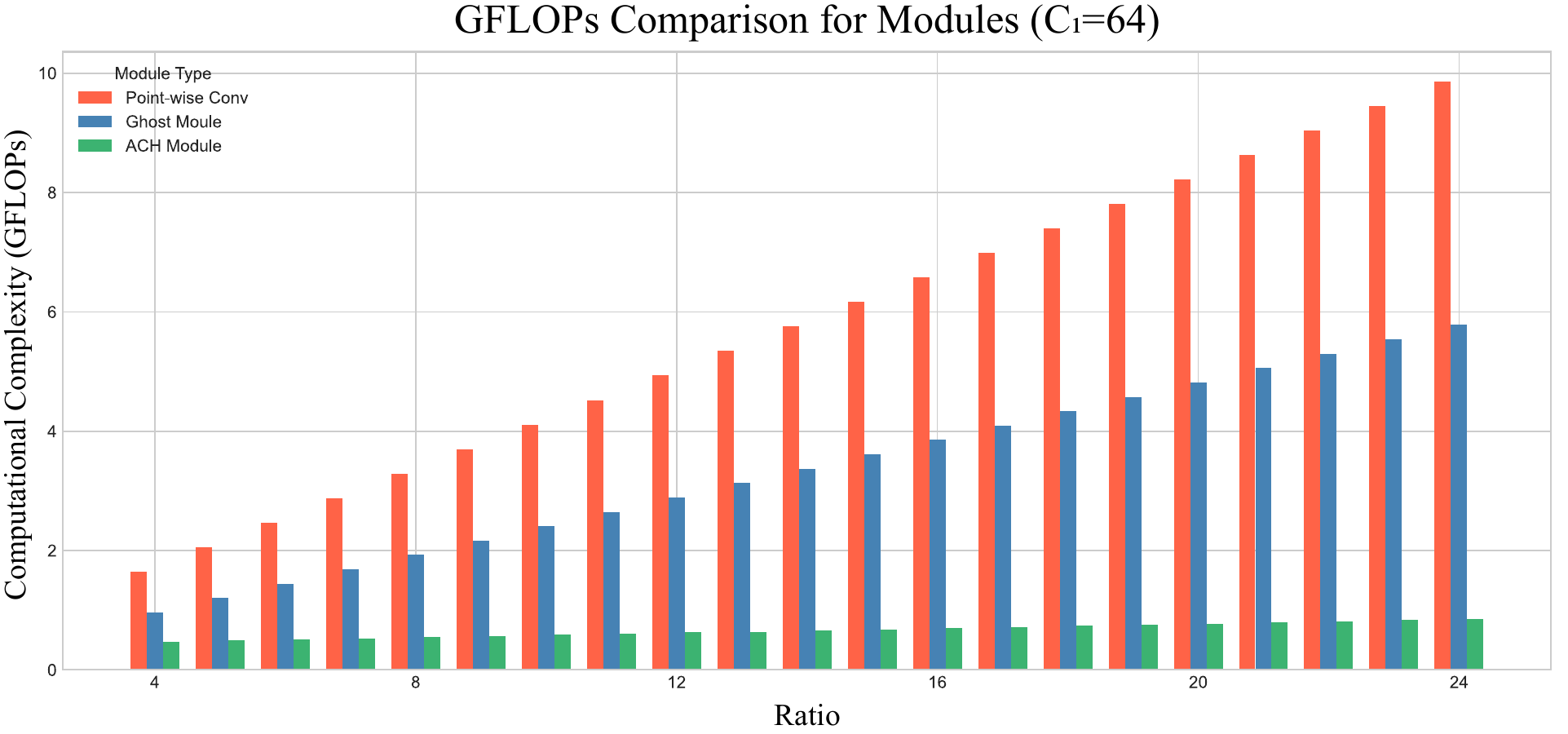}
\end{subfigure}
\hfill
\begin{subfigure}[t]{0.49\textwidth}
  \centering
  \includegraphics[width=\textwidth]{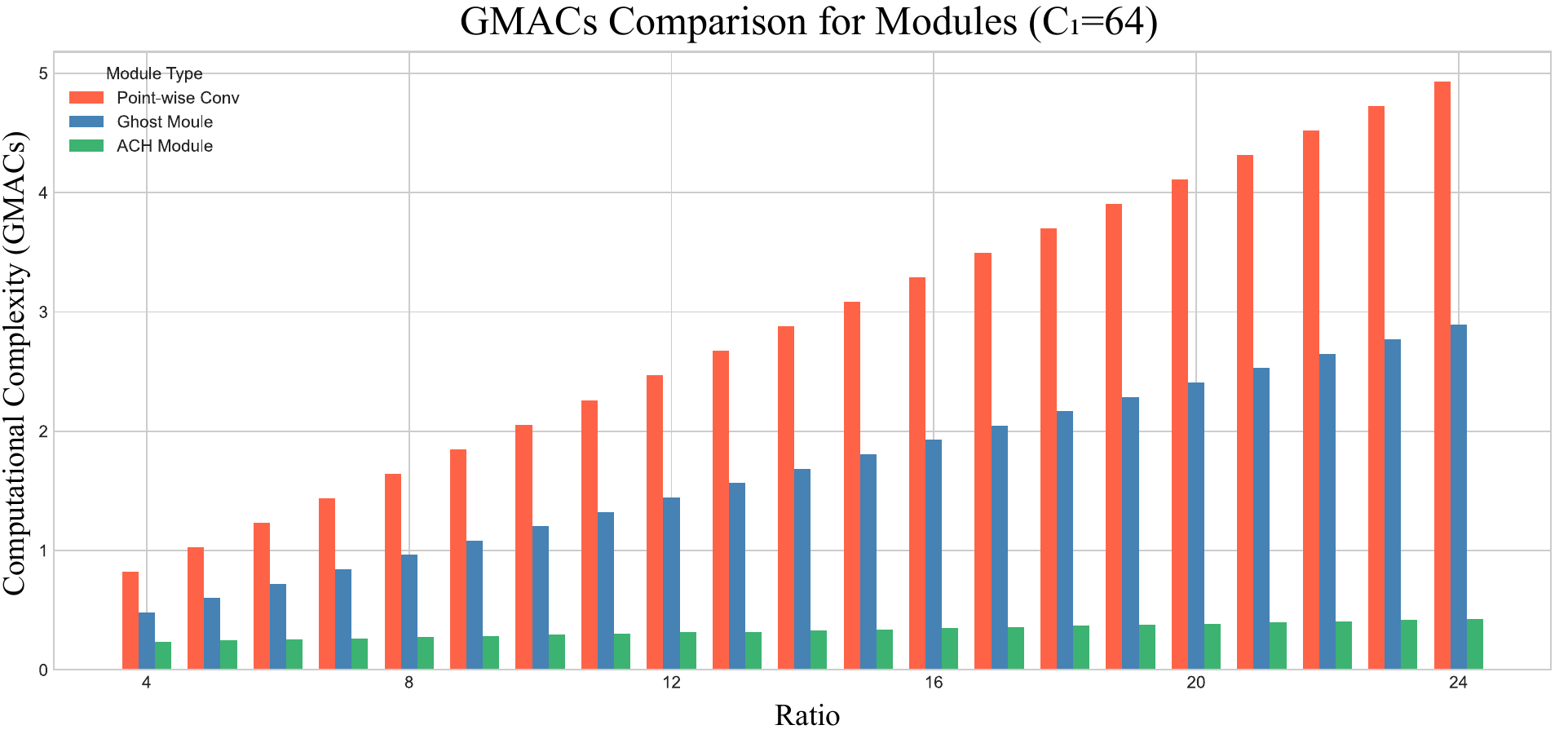}
\end{subfigure}
\caption{
   Comparison of computational efficiency under different expansion ratios
}
\label{fig:ratio_calc}
\end{figure}

The experiment is carried out for two specific situations: 
fixed 4-fold scaling ratio, 16-512 different input channel sizes; 
Fixed 64 input channel size, 4-24 times scaling ratio 
(224*224 per frame). We counted the Multiply-ACCumulate Operations 
(MACs) and Floating Point Operations (FLOPs) of the two groups of 
experiments as illustrated in \cref{fig:channel_calc,fig:ratio_calc}.

This ratio can be shown intuitively in the above experimental 
results. From the experimental results, the computational 
complexity ratio under different channel sizes is relatively 
fixed, while different scaling ratios, which is $m/n$, show a 
linear relationship.

Previous \cref{subsec:gpu} presented two GPU acceleration 
algorithms for ACH operators. One algorithm, the Direct-Indexing, 
is implemented as the name suggests. Another algorithm, 
the Parity-Balanced, could be written as \cref{alg:parity}.

\begin{figure}[h]
   \centering
   \includegraphics[width=0.8\textwidth]{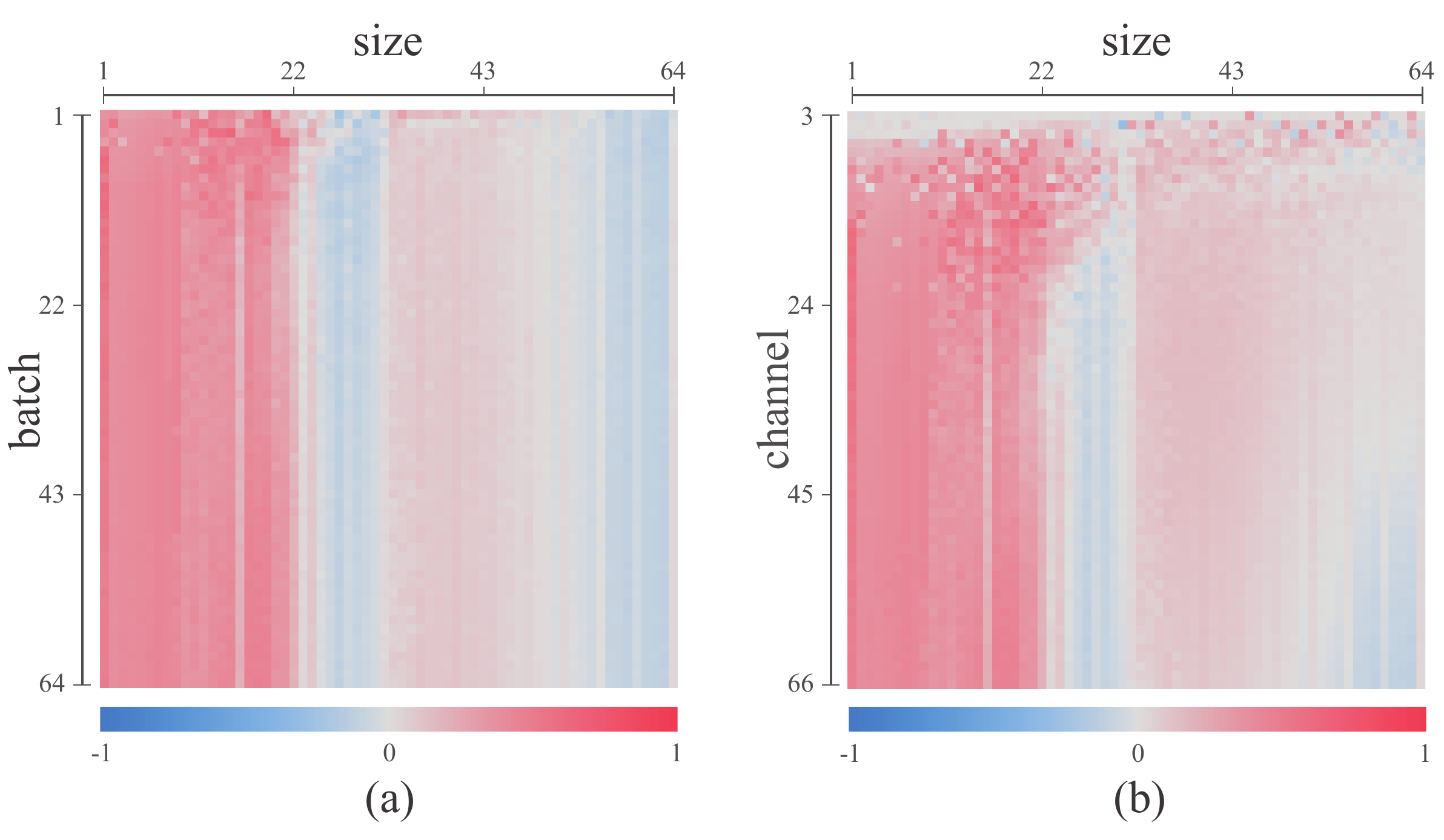}
   \caption{
      \textbf{Normalized difference heatmap of optimization 
      approaches runtime.}
      Color-coded visualization of relative performance between 
      direct-indexing ($A$) and parity-balanced ($B$) 
      approaches using $\frac{A-B}{A+B+\epsilon}$, 
      where red indicates A is slower (B more efficient) and 
      blue indicates the opposite. (a) Batch size versus 
      spatial dimensions scaling. (b) Channel count versus 
      spatial dimensions scaling.
   }
   \label{fig:heatmap}
\end{figure}

To systematically evaluate these methods under varying 
tensor configurations (batch/channel dimensions versus 
spatial sizes), we conducted comparative experiments 
using square matrices (same sized height \& width). 
See \cref{fig:heatmap} for the experiment details and results.

Both algorithms demonstrate relatively stable performance 
across varying batch sizes, indicating comparable parallelism 
in channel-agnostic scenarios. Despite both are expanding 
dimensionality, the parity-balanced approach exhibits 
superior optimization for high-channel tasks compared to 
high-batch scenarios, owing to its specialized load balancing 
for channel-dense tensors.

For feature maps with smaller spatial dimensions, 
the parity-balanced approach significantly outperforms 
direct-indexing due to:  
(1) The balanced approach's input tensor reuse pattern enhances 
L1/L2 cache hit rates in GPU global memory, reducing memory 
access latency while increasing arithmetic intensity per 
thread block through reduced thread block maintenance. 
(2) While appearing to introduce serialization, the balanced 
method effectively concentrates inevitable serial processes 
within individual thread blocks, as GPU core counts cannot 
simultaneously satisfy all computational demands for 
dimensionally dense small tensors, thereby avoiding 
context-switching overhead. 
(3) Direct-indexing requires separate thread block allocation 
per matrix computation, leading to underutilized warp resources 
when small matrices cannot fill the thread block size.

When spatial dimensions approach integer multiples of 32 
(thread block dimension), direct-indexing prevails due to 
thread blocks achieve near-saturation load conditions with peak 
artificial intensity, and the method's end-to-end processing 
better aligns with hardware scheduling optimizations.\\

\begin{algorithm}[H]
\caption{Parity-Balanced Indexing Strategy}\label{alg:parity}
\textbf{Input}: Channel count $c$\\
\textbf{Parameter}: Thread block group ID $id$\\
\textbf{Output}: Choosen channels $i, j$\\
\begin{algorithmic}[1]
\FOR{$it \gets 0$ to $c-1$}
   \IF{$it < id$ $\land$ $\neg((id - it) \bmod 2)$}
      \STATE $i \gets it$, $j \gets id$
   \ELSIF{$it > id$ $\land$ $(id - it) \bmod 2$}
      \STATE $i \gets id$, $j \gets it$
   \ELSE
      \STATE \textbf{continue}
   \ENDIF
   \STATE Compute Hadamard product for matrices $i$ and $j$
\ENDFOR
\end{algorithmic}
\end{algorithm}

\subsection{Extended Experiments}\label{supp:extended}


\begin{figure}[t]
   \centering
   \includegraphics[width=\textwidth]{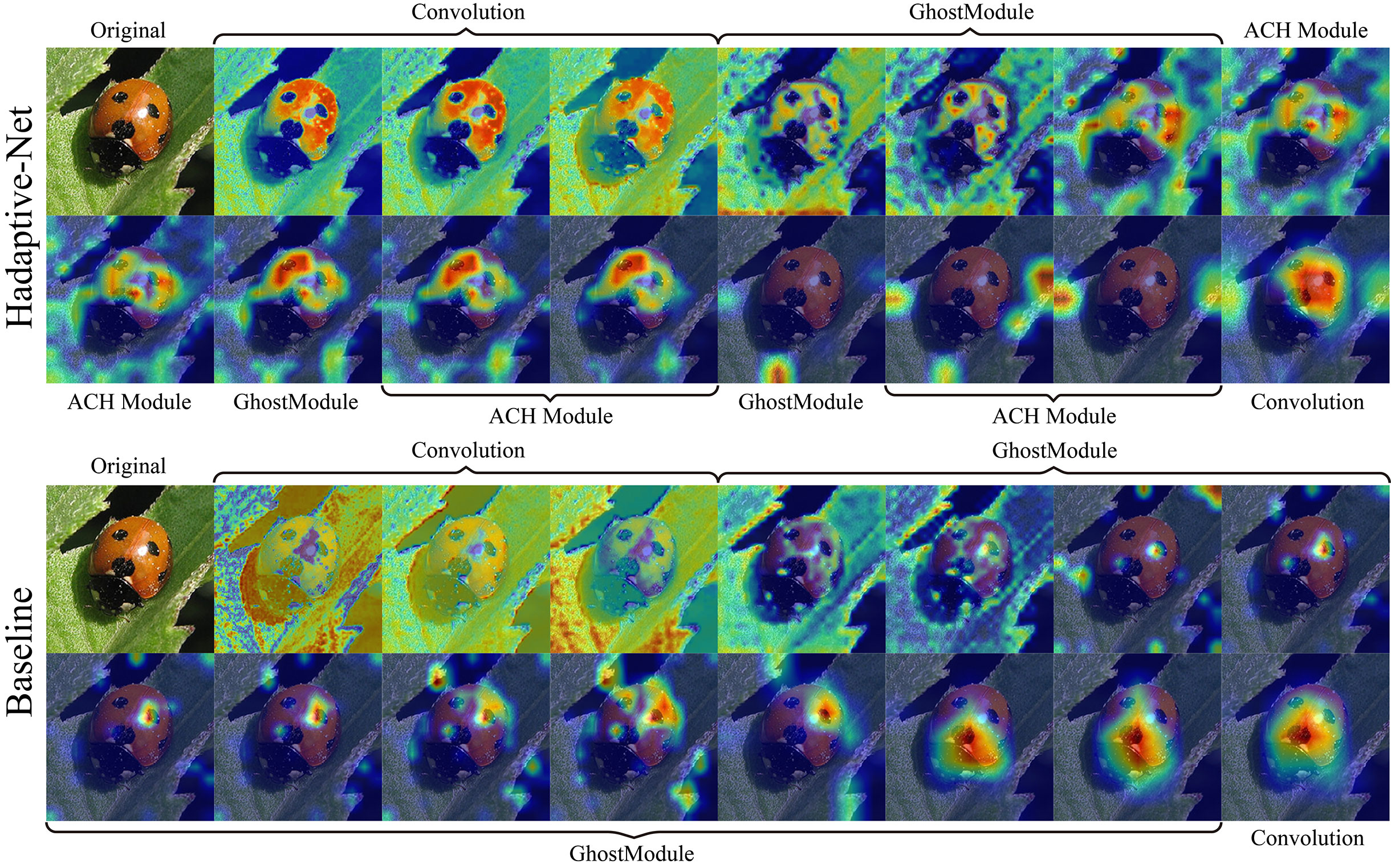}
   \caption{
      \textbf{Network visualization via Grad-CAM across layers (1).}
      Simple scenario: ladybug. Downward arrows denote downsampling layers.
   }
   \label{fig:gradcam1}
\end{figure}

\begin{figure}[t]
   \centering
   \includegraphics[width=\textwidth]{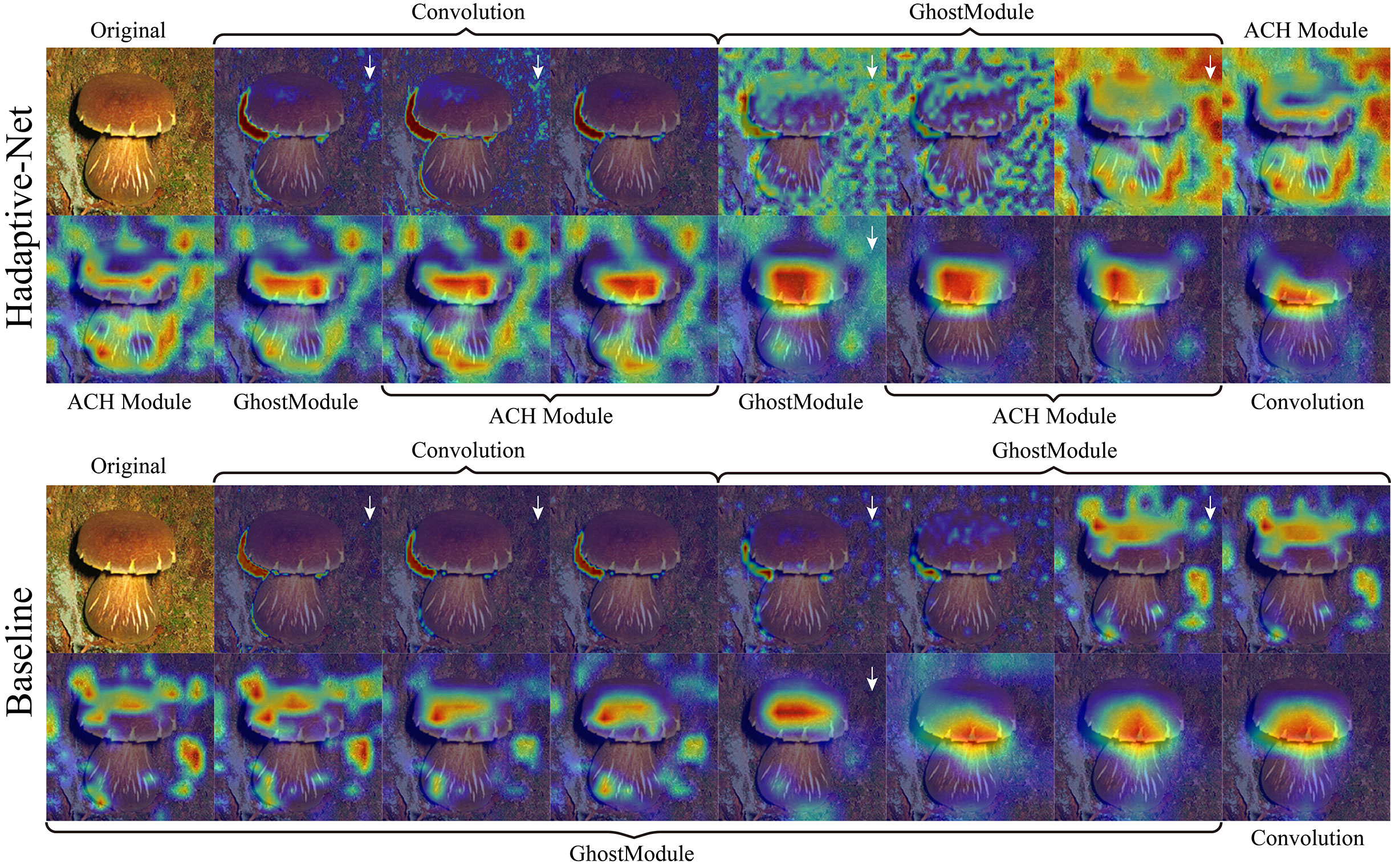}
   \caption{
      \textbf{Network visualization via Grad-CAM across layers (2).}
      Complex scenario: mushroom. Downward arrows denote downsampling layers.
   }
   \label{fig:gradcam2}
\end{figure}

\noindent\textbf{Grad-CAM}:
To further elucidate the role of the ACH module in enhancing 
the model’s representational capacity, we designed two sets of 
comparative experiments using Grad-CAM visualization to examine 
the changes brought by the ACH module compared to a conventional 
convolutional network. For clearer and more intuitive comparison, 
we adopted as the baseline a modified version of Hadaptive-Net-S 
in which all ACH modules were replaced with Ghost modules, in order 
to demonstrate the feature extraction pattern under purely linear 
transformations.

The first experiment, which is shown as \cref{fig:gradcam1}, involves a simple scenario, where a ladybug is 
clearly distinguishable from the background. The baseline model 
exhibits a standard processing pattern that progresses from texture 
analysis to focal emphasis. In contrast, Hadaptive-Net not only 
extracts texture more accurately, but also achieves target focus 
with fewer layers, while performing more precise edge segmentation. 
After the final downsampling step, the baseline model continues 
attempting to focus on the main subject, whereas Hadaptive-Net 
begins to attend to the edges of withered leaves, suggesting an 
attempt to capture higher-level semantic correlations.

The second experiment, which is shown as \cref{fig:gradcam2}, presents a more complex situation, where a 
mushroom exhibits some color overlap with the background. Compared 
to the baseline, Hadaptive-Net transitions more rapidly from the 
edge extraction phase to the target focusing phase, and explores a 
larger spatial area, indicating a larger effective receptive field.

In summary, the introduction of the ACH module not only reduces 
computational complexity but also endows the model with more 
powerful semantic representation capabilities.

\noindent\textbf{NLP Attempt}:
We tried to extend ACH module to NLP.
We conducted a comparative test on the 6-layers and 12-layers 
BERT \citep{DBLP:conf/naacl/DevlinCLT19}. 
The accuracy of the BERT model using different channel feature 
extractors (FFN, ACH module and standard FFN with unchanged 
middle layer dimension) was tested in SST-2 
\citep{DBLP:conf/emnlp/SocherPWCMNP13} and 10\% MNLI datasets 
\citep{DBLP:conf/naacl/WilliamsNB18}.

The models were evaluated on two standard natural language 
understanding benchmarks: the Stanford Sentiment Treebank 
(SST-2)  for binary sentiment classification and the 
Multi-Genre Natural Language Inference (MNLI) dataset for 
textual entailment. For SST-2, the model was trained and 
evaluated on the full dataset. To assess performance in a 
low-resource setting, the model was trained on a 
10\% stratified subset of the MNLI training set and evaluated 
on the full matched validation set.

The training configuration was consistent across both tasks. 
Models were trained for 3 epochs with a global batch size of 
32 and evaluated with a batch size of 64. The optimization 
used a learning rate of 2e-5 with a linear warmup over the 
first 10\% of the training steps and weight decay of 0.01. 
The models, which followed a BERT-base architecture 
(12 layers, 12 attention heads, 768-dimensional hidden states), 
were initialized with random weights. 
Input sequences were tokenized using the 
`bert-base-uncased` tokenizer with a maximum length of 128 
tokens. The sole evaluation metric was classification accuracy, 
calculated as the percentage of correctly predicted labels 
against the ground truth. All experiments were run on a 
single GPU without mixed-precision training.

As a result in \cref{tab:bert_results}, the performance of ACH module was not stunning 
enough to exceed the BERT baseline. However, from the perspective 
of the motivation of compressing the calculation scale, 
ACH module still plays a big role. Only adding a small amount of 
cross-Hadamard product in one step can approach the FFN without 
channel depth to a better level, revealing its potential as a 
unique algorithm in the field of NLP.

\begin{table}[!t]
\centering
\small
\caption{
   Replacements of ACH module on BERT.
}
\label{tab:bert_results}
\begin{tabular}{cccccc}
\toprule
Layers  & Modify              & In-Out-Channel & Mid-Channel & SST-2 & MNLI-10\% \\ \midrule
BERT-6  & FFN (fixed channel) & 764            & 764         & 64.7  & 36.7     \\
BERT-6  & ACH Module          & 764            & 2043        & 76.2  & 41.8     \\
BERT-6  & FFN                 & 764            & 2048        & 80.3  & 46.7     \\ \midrule
BERT-12 & FFN (fixed channel) & 764            & 764         & 53.1  & 34.3     \\
BERT-12 & ACH Module          & 764            & 2043        & 75.9  & 41.2     \\
BERT-12 & FFN                 & 764            & 2048        & 80.6  & 45.5     \\ \bottomrule
\end{tabular}
\end{table}

As we are mainly engaged in CV related work and lack relevant 
experience in NLP field, the experimental setup may be a little 
rough. If there are any problems, readers are welcome to correct 
them.

\section{The Use of Large Language Models}

In the preparation of this work, the author(s) utilized a 
Large Language Model (LLM) primarily to aid in polishing and 
refining the writing. The tool was used for purposes such as 
improving grammatical correctness, enhancing sentence fluency, 
and rephrasing for clarity. All ideas, theoretical analyses, 
experimental designs, results, and conclusions remain entirely 
those of the author(s). The final manuscript has been thoroughly 
reviewed and edited by the author(s), who take full responsibility 
for all content presented herein.

\end{document}